\documentclass[10pt,twocolumn,letterpaper]{article}

\usepackage[pagenumbers]{wacv} 

\usepackage{algpseudocode}
\usepackage[ruled,vlined]{algorithm2e}
\usepackage{amsfonts}
\usepackage{amsmath}
\usepackage{amssymb}
\usepackage{bbm}
\usepackage{caption}
\usepackage{kotex} 
\usepackage{lipsum}
\usepackage{multirow}
\usepackage{pifont}
\usepackage{subcaption}
\usepackage{setspace} 
\usepackage{wrapfig}
\usepackage[dvipsnames]{xcolor}
\usepackage{orcidlink}
\usepackage{enumitem}
\definecolor{cgood}{HTML}{b0a92a}

\DeclareSymbolFont{extraup}{U}{zavm}{m}{n}
\DeclareMathSymbol{\varheart}{\mathalpha}{extraup}{86}
\DeclareMathSymbol{\vardiamond}{\mathalpha}{extraup}{87}

\newcommand{\bgcolor}[2]{\setlength{\fboxsep}{0pt}\colorbox{#1}{\strut #2}}
\newcommand{\BGcolor}[3][HTML]{\definecolor{mycolor}{HTML}{#2}\bgcolor{mycolor}{#3}}

\newcommand{\metrictablefirst}[1]{{\BGcolor{FFB2B3}{#1}}}
\newcommand{\metrictablesecond}[1]{{\BGcolor{FFDAB2}{#1}}}
\newcommand{\metrictablethird}[1]{{\BGcolor{FFFFB3}{#1}}}

\graphicspath{{images/}}

\makeatletter
\renewcommand{\fnum@figure}{Fig. \thefigure}
\makeatother

\makeatletter
\renewcommand{\fnum@table}{Table \thetable}
\makeatother

\usepackage{titletoc}

\newcommand{\worldwideweb}{\raisebox{-1.5pt}{\includegraphics[height=1.05em]{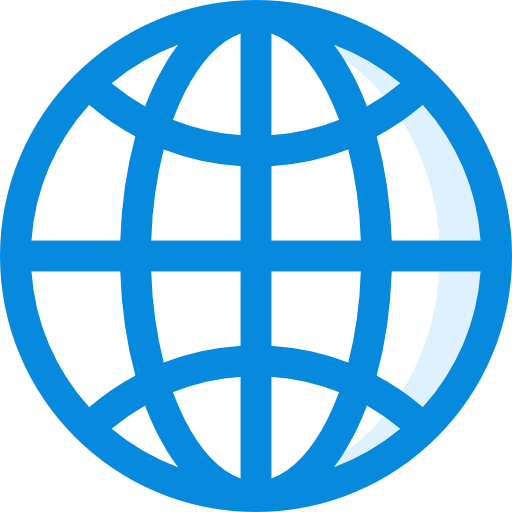}}\xspace}
\newcommand{\github}{\raisebox{-1.5pt}{\includegraphics[height=1.05em]{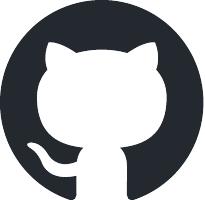}}\xspace}
\definecolor{wacvblue}{rgb}{0.21,0.49,0.74}

\def\wacvPaperID{925} 
\def\confName{WACV}
\def\confYear{2027}

\title{Difix3D-W: Distractor-Free Few-Shot 3D Gaussian Splatting in the Wild}

\author{
    Wongi Park$^{1}$ \quad
    Jordan A. James$^{2}$ \quad
    Myeongseok Nam$^{3}$ \quad
    Minjae Lee$^{4}$ \quad \\
    Soomok Lee\textsuperscript{\dag}$^{5}$ \quad
    SangHyun Lee\textsuperscript{\dag}$^{1}$ \quad
    William J. Beksi\textsuperscript{\dag}$^{2}$ 
    \vspace{2mm}
\\
 \textsuperscript{1}Ajou Univerity \quad
 \textsuperscript{2}University of Texas at Arlington \quad
 \textsuperscript{3}GenGenAI \quad \\
 \textsuperscript{4}Seoul National University  \quad 
 \textsuperscript{5}Kennesaw State University 
\\
\small 
   \texttt{\{psboys, sanghyunlee\}@ajou.ac.kr} \;
   \texttt{slee337@kennesaw.edu} \;
   \texttt{william.beksi@uta.edu}
   \\
   {\github \href{https://github.com/robotic-vision-lab/Diffix3D-W}{{\text{Code}}}}
\quad \quad
{\worldwideweb \href{https://robotic-vision-lab.github.io/Diffix3D-W/}{{\text{Project page}}}}
}

\begin{document}

\twocolumn[{%
\renewcommand\twocolumn[1][]{#1}%
\maketitle
\begin{center}
    \centering
    \captionsetup{type=figure}
    \includegraphics[width=\textwidth]{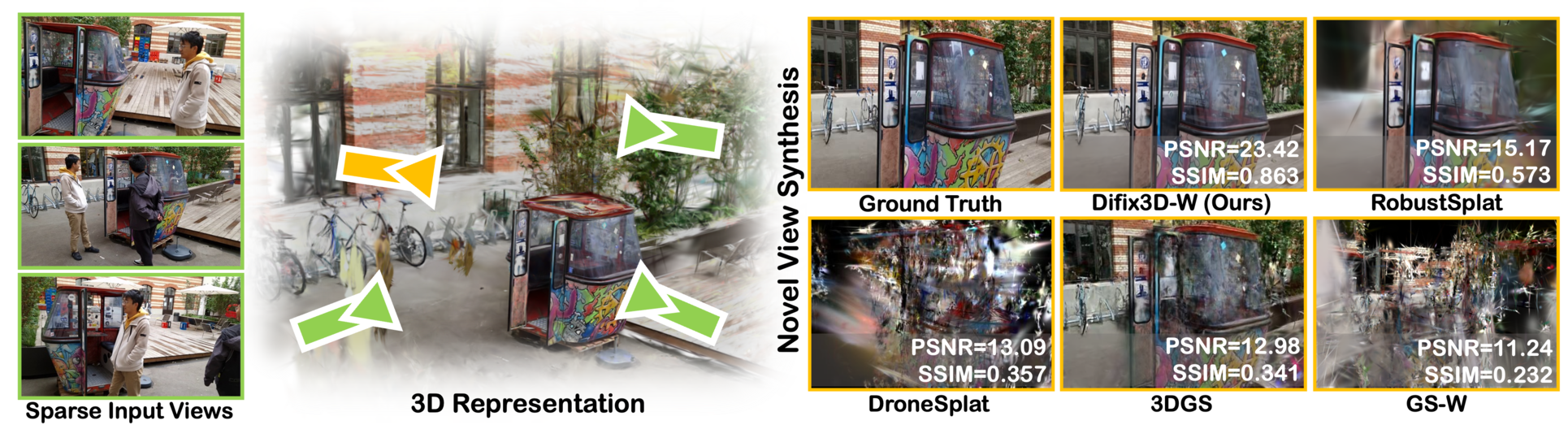}
    \vspace{-2em}
    \caption{Given a sparse set of images, our approach effectively renders 3D novel view synthesis in real-world scenarios, including distractors. To the best of our knowledge, we are the first to address 3D novel view synthesis from a sparse set of unconstrained real-world images.}
    \label{fig:overview}
\end{center}
}]

\footnotetext[1]{\textsuperscript{\textdagger}Corresponding author.}

\begin{abstract}
We propose Difix3D-W, a 3D novel sparse-view synthesis framework for unconstrained real-world scenarios that contain distractors, occlusion, and appearance variation. Unlike existing methods that primarily perform novel-view synthesis from a sparse set of constrained images without transient elements or leverage unconstrained dense image collections in real-world scenarios, our method utilize sparse unconstrained images, showing high-quality 3D rendering results. To do this, we introduce reference-guided view refinement with a redesigned one-step diffusion model using a transient mask and a reference image to mitigate artifacts in rendered views, enhancing the 3D representation in the Gaussian field. Furthermore, we address sparse regions in the Gaussian field leveraging sparsity-aware Gaussian replication strategy to amplify Gaussians in the sparse regions and alleviate deficient camera viewpoint issues. Finally, we utilize LoRA and regularization to maintain 3D multi-view consistency. Extensive experiments demonstrate that our method consistently outperforms existing methods. This advancement paves the way for realizing real-world scenarios without labor-intensive data acquisition.
\end{abstract}
\vspace{-1em}
\section{Introduction}
\label{sec:introduction}
Rendering a realistic representation of a 3D scene from a collection of images
is a fundamental challenge in computer vision and graphics. Solving this problem
is crucial for applications such as
robotics~\cite{huang2026enerverse,huang2025gaussiannexus},
autonomous
driving~\cite{gan2025gaussianocc,zhao2025drivedreamer4d,wei2026parkgaussian}, VR/AR~\cite{huang2026enerverse,shen2025gaussianshopvr,tu2025vrsplat}, and 3D
content generation~\cite{ling2024align,ren2024l4gm,ren2025gen3c}. While
contemporary methods~\cite{chen2024hac,lu2024scaffold,chen2025hac++,wu2025bg}
excel at reconstructing 3D scenes from a dense sets of images, they struggle with
sparse sets due to limited geometric information, depth ambiguity, and deficient
perspectives.

\begin{figure*}[!t]
\centering
\includegraphics[width=\textwidth]{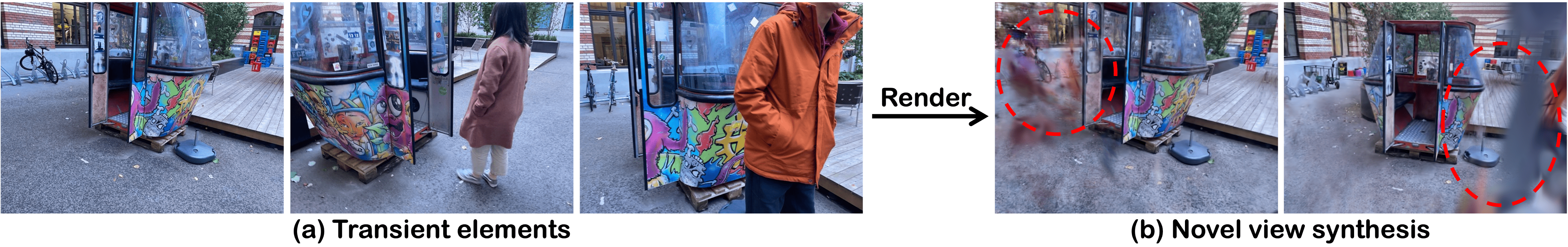}
\vspace{-2em}
\caption{We observe that (a) distractors are ephemeral, appearing across diverse regions, and (b) using real-world images for 3D novel view synthesis results in corrupted images. These observations provide a direction for improving corrupted images by leveraging
other views.} 
\label{fig:motivation}
\vspace{-1.5em}
\end{figure*}

\noindent \textbf{Prior work.} To overcome these challenges, several approaches
have been proposed, which generally fall into three paradigms: (i) multi-stage
training-based methods~\cite{yin2024fewviewgs,dai2025eap,zhao2025self} that
handle unobserved areas by increasing training time with multi training
stages; (ii) depth regularization-based
techniques~\cite{li2024dngaussian,zhu2024fsgs,kong2025generative} that leverage
foundation models (\eg, DepthAnything~\cite{yang2024depth},
DPT~\cite{ranftl2021vision}) to regularize depth; (iii) diffusion-based
methods~\cite{xiong2023sparsegs,wu2024reconfusion,cheng2025perspective,paliwal2025ri3d,yin2025gsfixer}
that refine rendered views or generate unobserved viewpoints to distill a 3D representation.

\noindent \textbf{Challenges.} Prior works have shown that high-quality 3D
rendering can be realized from a sparse set of constrained images. However,
these approaches fail to utilize unconstrained real-world scenarios due to the
presence of distractors, appearance variations, and dynamic objects
(Fig.~\ref{fig:overview}). Several
methods~\cite{martin2021nerf,kulhanek2024wildgaussians,tang2024nexussplats,zhang2024gaussian,park2026forestsplats}
attempt to address this issue by training vision foundation models (\eg,
SAM~\cite{kirillov2023segment}, DINO~\cite{oquab2023dinov2},
Diffusion~\cite{tang2023emergent}). Nonetheless, these techniques struggle to
identify distractors due to the limited number of images, making it difficult to
generalize.

\noindent \textbf{Motivation.} Inspired to solve these challenging issues, we
observe that leveraging a sparse set of real-world images leads to
multi-view inconsistency and struggles to capture distractors. In particular, we analyze how a sparse set of real-world images can fail
to provide high-quality results in 3D novel view synthesis
(Fig.~\ref{fig:motivation}). Furthermore, we note that merely leveraging
positional gradients to mitigate sparsity leads to artifact issues in the rendering results.

\noindent \textbf{Solution.} To address this dilemma, we introduce Difix3D-W, a
framework that enables 3D novel view synthesis from a sparse set of real-world
images with distractors. Specifically, we propose a reference-guided view
refinement by utilizing a redesigned diffusion model to refine rendered views
using a reference view and a transient mask. Moreover, to tackle sparsity in the Gaussian field, we amplify Gaussians in the sparse regions to construct a dense Gaussian field. We also employ low-rank adaptation (LoRA) and the score distillation sampling (SDS) loss to prevent model collapse and maintain geometric consistency. Extensive experiments on various scenarios (\eg,
NeRF On-the-go~\cite{ren2024nerf}, Photo Tourism~\cite{snavely2006photo},
LLFF~\cite{mildenhall2021nerf}) demonstrate that Difix3D-W outperforms existing
methods. 

\noindent \textbf{Key distinction.} To the best of our knowledge, Difix3D-W is
the first framework to tackle 3D reconstruction from a sparse set of real-world
images that include diverse distractors. This is not an incremental extension of
existing methods. Notably, Difix3D-W (i) bridges the gap between a sparse set of images and diverse real-world scenarios with distractors, (ii)
enables high-quality 3D rendering results without sacrificing significant time complexity, and (iii) can be utilized in unconstrained/constrained scenarios via plug-and-play. In summary, our contributions are as follows.
\begin{itemize}[leftmargin=*]
\setlength{\parsep}{0pt}
\setlength{\parskip}{0pt}
  \item \textbf{Impact.} We propose a novel framework for sparse-view
  synthesis in unconstrained scenarios with distractors,
  without significantly increasing time complexity. 
  \item \textbf{Versatility.} Our method can be utilized through plug-and-play
  with a redesigned one-step diffusion model in constrained or unconstrained scenarios.
  \item \textbf{Effectiveness.} Extensive experiments demonstrate that Difix3D-W
  outperforms the prior methods (e.g., PSNR
  -\textcolor{cgood}{11.2\%$\uparrow$}, SSIM - \textcolor{cgood}{10.5\%$
  \uparrow$}, LPIPS - \textcolor{cgood}{4.3\%$ \uparrow$}).
\end{itemize}
\begin{figure*}[!t]
\centering
\includegraphics[width=\textwidth]{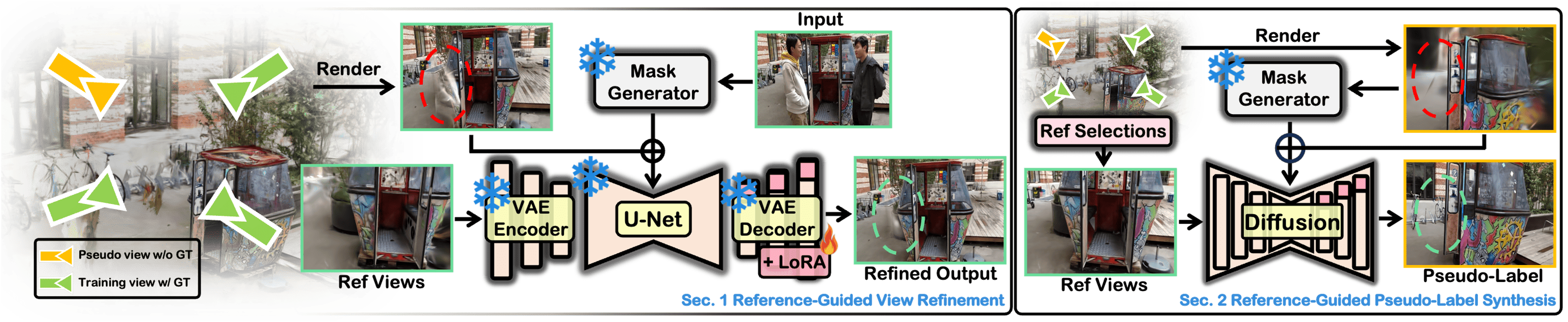}
\vspace{-2.2em} \caption{An overview of Difix3D-W. We refine rendered images
utilizing a reference view and a transient mask using a one-step DM without
significantly increasing time complexity. Moreover, to tackle deficient perspectives, we
generate a pseudo-view to distill the 3D representation and improve
consistency.} 
\label{fig:method}
\vspace{-1.3em}
\end{figure*}
\section{Related Work}
\noindent \textbf{Dense-view synthesis for unconstrained scenarios.} Recently,
neural radiance fields (NeRFs)~\cite{mildenhall2021nerf} and 3D Gaussian
splatting (3DGS)~\cite{kerbl20233d} have been widely used in 3D reconstruction.
Prior works fall into three paradigms: (i) residual-based
approaches~\cite{chen2022hallucinated,li2023nerf,zhang2024gaussian,park2026forestsplats}
that only leverage photometric error to identify transient elements; (ii) semantic-based methods~\cite{kulhanek2024wildgaussians,
sabour2024spotlesssplats,xu2024wild} that utilize semantic features obtained
from a foundation model (\eg, SAM~\cite{kirillov2023segment},
DINO~\cite{oquab2023dinov2}, Diffusion~\cite{tang2023emergent}) to generate
transient masks; (iii) heuristic-based techniques
\cite{bao2024distractor,chen2024nerf,tang2025dronesplat} that exploit 2D masks
from SAM to construct transient masks. Although existing methods show impressive
results, they rely on a dense image collection which necessitates a large amount
of time to collect. In contrast, our approach handles a sparse set of images
that include distractors for 3D novel view synthesis.

\noindent \textbf{Sparse-view synthesis.} While
previous works~\cite{lu2024scaffold,chen2024hac,chen2025hac++,wu2025bg} utilize
dense image collections without transient elements, recent approaches make an
effort to achieve 3D reconstruction from a sparse set of images. These
techniques can be separated into the following: (i) multi-stage training-based
methods~\cite{yin2024fewviewgs,dai2025eap,zhao2025self} that render 3D views via
diverse training strategies; (ii) depth-based approaches~\cite{li2024dngaussian,
zhu2024fsgs,kong2025generative} that leverage a foundation model (\eg, Depth
Anything~\cite{yang2024depth}, DPT~\cite{ranftl2021vision}) to regularize depth; (iii) diffusion-based
methods~\cite{xiong2023sparsegs,wu2024reconfusion,cheng2025perspective,paliwal2025ri3d,yin2025gsfixer}
that refine rendered views and generate pseudo-views to handle deficient camera
viewpoints. Nevertheless, these methodologies are difficult to utilize in
real-world scenarios due to distractors. Several works
\cite{li2025ms,li2025sparsegs,zhang2025rgs} try to address sparse-view
synthesis, yet these techniques only focus on appearance variation and not on
substantial occlusions. In contrast, our method can be used with sparse
views that include diverse distractors.

\noindent \textbf{Adaptive density control.} An adaptive density control (ADC)
strategy in 3DGS works via two operations: pruning and densification. These
actions fill sparse regions and add fine details. Recent
works~\cite{zhang2024pixelgs,afane2025atp,xu2025ad} aim to enhance the Gaussian
fields for various purposes including (i) improving memory
efficiency~\cite{niedermayr2024compressed,lee2024compact}, (ii) filling in
deficient regions~\cite{xiong2023sparsegs,chung2024depth}, and (iii) enhancing
rendering quality and consistency~\cite{zhang2024pixelgs,zeng2025frequency}.
Nonetheless, these approaches focus on constrained images, which makes them
difficult to utilize in real-world settings. Several
methods~\cite{fu2025robustsplat,park2026forestsplats} address ADC in
unconstrained scenarios to align the Gaussians, yet they are not easy to use
with a sparse image collection due to a lack of consideration for sparsity in
the Gaussian field. Specifically, ForestSplats~\cite{park2026forestsplats}
introduced uncertainty primitives to effectively align the Gaussians.
Conversely, in this work we amplify the Gaussians to mitigate the sparsity issue and address deficient perspectives.

\section{Preliminaries}
\vspace{0.2\baselineskip}
\noindent \textbf{3D Gaussian splatting.} 3DGS represents a scene as a
set of anisotropic Gaussians $\mathcal{G} = \{g_k\}_{k=1}^N$. Each
$\mathcal{G}$ is parameterized by a position $\mu_k \in \mathbb{R}^{3}$, a
covariance matrix $\Sigma_k \in \mathbb{R}^{3\times3}$ decomposed into a scaling
$S_k \in \mathbb{R}^{3}$, a rotation matrix $R_k \in \text{SO(3)}$, an opacity
parameter $\alpha_k \in [0, 1]$, and view-dependent colors $c_k \in
\mathcal{C}^{N_{\text{SH}}}$ represented via spherical harmonic (SH)
coefficients $N_{\text{SH}}$. The color $\hat{C}$
of a pixel can be computed by blending ordered Gaussians overlapping the pixel,
\begin{equation}
  \hat{C}=\sum\limits_{k=1}^{N} c_{k} \alpha_{k} \prod_{j=1}^{k-1}\left(1-\alpha_j\right).
\end{equation}
The attributes of the Gaussians $\{g_k\}_{k=1}^N$ are optimized by minimizing
the photometric loss between the rendered image $\hat{I} \in \mathbb{R}^{3
\times H \times W}$ and ground-truth (GT) image $I_{\text{GT}} \in \mathbb{R}^{3
\times H \times W}$, 
\begin{equation}
  \mathcal{L}_{\mathrm{GS}} = (1-\lambda) \mathcal{L}_1(\hat{I}, I_{\text{GT}})+\lambda \mathcal{L}_{\text {D-SSIM}}(\hat{I}, I_{\text{GT}}),
\end{equation}
where $\mathcal{L}_{1}$ is the $L_{1}$ loss, $\mathcal{L}_{\text{D-SSIM}}$ is
the SSIM loss, and $\lambda$ is a weighting factor.

\noindent \textbf{Diffusion models.} A diffusion model (DM) generates an image
by progressively denoising Gaussian noise $\epsilon \sim \mathcal{N}(0, I)$.
Specifically, in the forward process noise is injected into the clean data
$\mathbf{x}_0 \in \mathbb{R}^{n\times3\times h \times w}$, promoting a sequence
of increasingly noisy data $\mathbf{x}_0, \ldots, \mathbf{x}_t$. The reverse
procedure then utilizes the DM to invert this process by iteratively denoising
back from $\mathbf{x}_t$ to reconstruct $\mathbf{x}_0$. The noise predictor
$\epsilon_\theta$ is optimized using a denoising objective,
\begin{equation}
  \min _\theta \mathbb{E}_{t \sim \mathcal{U}(0,1), \epsilon \sim \mathcal{N}(\mathbf{0}, \boldsymbol{I})}\left[\left\|\epsilon_\theta\left(\boldsymbol{x}_t; \tau, t\right)-\epsilon\right\|_2^2\right],
\end{equation} 
where $\tau$ represents an optional conditioning prompt (\eg, an image context
or text prompt) and $\mathcal{U}(0,1)$ denotes a uniform distribution. Following
prior work, we utilize a fixed discretization with the diffusion time $t$ drawn
from a uniform distribution over $[0, 1000]$. The maximum diffusion time is
selected to guarantee that the data is entirely transformed into Gaussian noise.
\section{Method}
Inspired by the preceding observations, we propose a comprehensive framework
that leverages 3DGS for novel view synthesis from a sparse set of unconstrained
images. As illustrated in Fig.~\ref{fig:method}, Difix3D-W consists of two key
components: (i) reference-guided view refinement, which employs a redesigned DM to refine
rendered views by leveraging a reference view and a transient mask
(Sec.~\ref{subsec:reference-guided_view_refinement}); (ii) pseudo-label
synthesis that addresses sparse camera viewpoints and solves the issue of sparsity in the Gaussian field via
amplification (Sec.~\ref{subsec:reference-guided_pseudosynthesis}). Finally, to maintain 3D
consistency, we introduce regularization and optimization to mitigate collapse
and artifact issues (Sec.~\ref{subsec:regularization_optimization}).

\subsection{Reference-Guided View Refinement}
\label{subsec:reference-guided_view_refinement}
\textbf{Mask generator.} To capture distractors given a sparse set of images,
prior methods use semantic features obtained from a frozen DINOv2~\cite{oquab2023dinov2} to construct transient
masks, 
\begin{equation}
  \mathcal{M}_{t} =  \mathrm{Sigmoid}\!\left(\psi_{\theta}\!\left(f(I_{\text{GT}})\right)\right), 
\end{equation}
where $\psi_{\theta}$ and $f$ denote learnable MLP layers and the DINOv2 feature
extractor. To optimize MLP layers, existing approaches~\cite{kulhanek2024wildgaussians,sabour2024spotlesssplats,
fu2025robustsplat} use similarity maps and photometric loss to train MLP layers.
However, training with a sparse set of images may fail to capture transient
elements due to limited examples. To fix this issue, we utilize Grounded SAM~\cite{ren2024grounded} to capture distractors.
Specifically, transient masks are generated using a ground-truth image and text
description $T_{\mathrm{text}}$ as input, 
\begin{equation}
  \mathcal{M}_{t} = \mathcal{S}_{\theta}(I_{\text{GT}}, T_{\text{text}}).
\end{equation} 
By leveraging them, in contrast to existing methods~\cite{kulhanek2024wildgaussians,
sabour2024spotlesssplats,fu2025robustsplat}, we identify transient elements, achieving generalization regardless of
the number of samples.



\begin{figure}[!t]
\centering
\includegraphics[width=\linewidth]{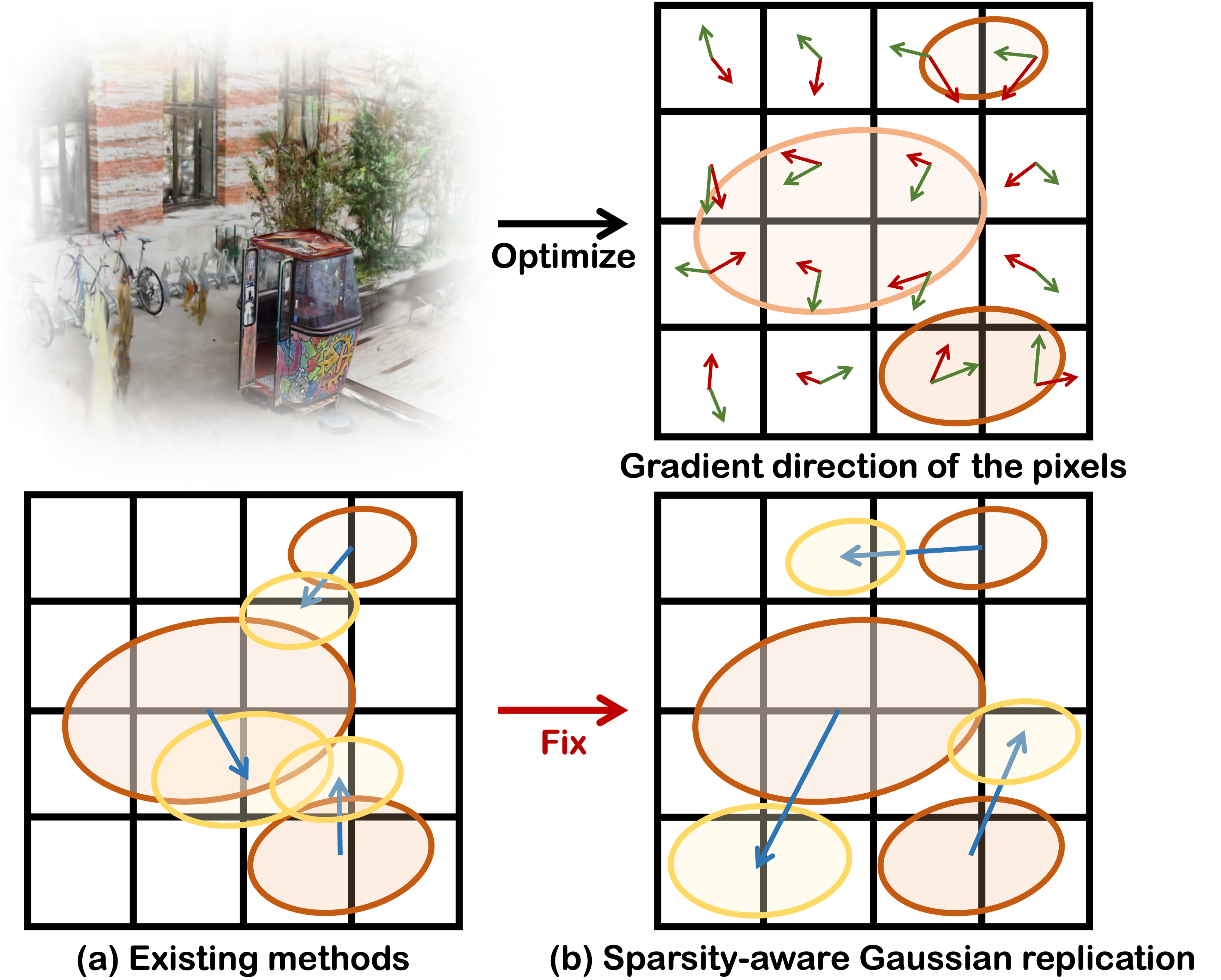}
\vspace{-1.7em}
\caption{An overview of sparsity-aware Gaussian replication. (a) Existing work
densifies the Gaussians leveraging positional gradients, which makes it
difficult to address sparsity in the Gaussian field. (b) In contrast, we utilize
opacity information to fill sparse regions, thus solving issue of sparsity for insufficient camera viewpoints.} 
\label{fig:sagr}
\vspace{-1.3em}
\end{figure}

\noindent \textbf{Rendered view refinement.} Previous works~\cite{liu20243dgs,wu2024reconfusion,cheng2025perspective,paliwal2025ri3d} primarily enhance rendered views by employing reference views without distractors. However, leveraging prior approaches is extremely
difficult in unconstrained real-world scenarios due to transient elements. To tackle this issue, we redesign Difix3D+~\cite{wu2025difix3d+}, a one-step DM, to
utilize a reference view and a transient mask in order to refine the rendered views. Unlike prior
work~\cite{wu2024reconfusion,paliwal2025ri3d,wu2025difix3d+}, we use a corrupted
rendered image $\hat{I}$ and a reference view $\hat{I}_{\text{ref}}$ to generate
a refined image via 
\begin{equation}
   \tilde{I}_{\text{ref}} = D_{\theta}(\hat{I}, \hat{I}_{\text{ref}}, \mathcal{M}_{t}),
   \label{eq:diffusion_model}
\end{equation}
where $D_{\theta}$ denotes the DM. Concretely, we modify the cross-attention in
the DM to selectively refine corrupted regions utilizing transient masks. In
each cross‑attention layer of the denoising U‑Net, we compute the attention map
between the query $\ell_{Q}$ derived from a rendered image $\hat{I}$, key
$\ell_{K}$, and value $\ell_{V}$ obtained from a reference image
$\hat{I}_{\text{ref}}$ as 
\begin{equation}
  A = \mathcal{M}_{t} \odot \mathrm{softmax}\left(\frac{\ell_{Q}\ell_{K}^{\top}}{\sqrt{d}}\right)\ell_{V} + (1-\mathcal{M}_{t}) \odot A_{\mathrm{self}},
  \label{eq:cross-attention}
\end{equation}
where $A_{\mathrm{self}}$ is obtained by a self-attention mechanism without the
reference view and $d$ denotes the feature dimension of $\ell_{K}$. Employing
the cross-attention mechanism ensures that reference views refine the masked
regions in the rendered image, while the rest of the image remains largely
unaffected in the rendered views. Finally, refined rendered views are utilized to optimize 3D Gaussians by redefining
$\mathcal{L}_{\text{GS}}$ as 
\begin{equation}
  \mathcal{L}_{\text{photo}} =  (1-\lambda) \mathcal{L}_1(\hat{I}, \tilde{I}_{ref})+\lambda \mathcal{L}_{\text {D-SSIM}}(\hat{I}, \tilde{I}_{ref}).
\end{equation} 
By leveraging the reference-guided view refinement, we mitigate artifact issues in rendered views despite the presence of distractors and without increasing the training time complexity. Note that we randomly select reference views from camera viewpoints with available GT.
\begin{table*}[!t]
	\centering
    \renewcommand{\arraystretch}{1}   
	\resizebox{\linewidth}{!}{
		\begin{tabular}{cccccccccccccc}
			\toprule
			\multirow{2}{*}{Method} &\multirow{2}{*}{GPU hrs / FPS}& \multicolumn{3}{c}{3-view} &  \multicolumn{3}{c}{6-view}  & \multicolumn{3}{c}{9-view} & \multicolumn{3}{c}{Average} \\
			&  & { PSNR$\uparrow$} & { SSIM$\uparrow$} & { LPIPS$\downarrow$} & { PSNR$\uparrow$} & { SSIM$\uparrow$} & { LPIPS$\downarrow$} & { PSNR$\uparrow$} & { SSIM$\uparrow$} & { LPIPS$\downarrow$} & { PSNR$\uparrow$} & { SSIM$\uparrow$} & { LPIPS$\downarrow$} \\
                        \midrule
            \midrule
			3DGS \cite{kerbl20233d} & 0.8 / 103 & 11.67 & 0.273 & 0.562 & 12.90 & 0.309 & 0.521 & 14.28 & 0.394 & 0.479 & 12.95 & 0.325 & 0.521 \\
            RobustSplat \cite{fu2025robustsplat} & 0.72 / 104 & 12.14 & 0.305 & 0.628 & 13.70 & 0.425 & 0.587 & 11.84 & 0.291 & 0.638 & 12.56 & 0.340 & 0.617 \\
            GS-W \cite{zhang2024gaussian} & 3.2 / 85 & 11.16 & 0.296 & 0.627 & 13.63 & 0.370 & 0.566 & 13.37 & 0.370 & 0.515 & 12.72 & 0.345 & 0.569 \\
            DroneSplat \cite{fu2025robustsplat} & 0.63 / 113 & 11.64 & 0.249 & 0.555 & 13.40 & 0.351 & \metrictablethird{0.491} & 14.12 & 0.364 & \metrictablethird{0.449} & 13.05 & 0.345 & 0.522 \\
            WildGaussians \cite{kulhanek2024wildgaussians} & 1.5 / 107 & \metrictablethird{13.43} & \metrictablethird{0.423} & \metrictablethird{0.499}	& \metrictablethird{14.25}	& \metrictablethird{0.484} & 0.494 & \metrictablethird{14.57} & \metrictablethird{0.486} & 0.469 & \metrictablethird{13.48} & \metrictablethird{0.406} & \metrictablethird{0.506} \\
            Difix3D+ \cite{wu2025difix3d+} & 2.5 / 103 & \metrictablesecond{15.83} & \metrictablesecond{0.508} & \metrictablesecond{0.481}	& \metrictablesecond{16.34} & \metrictablesecond{0.558} &\metrictablesecond{0.460} & \metrictablesecond{16.86}	& \metrictablesecond{0.580} & \metrictablesecond{0.387}	& \metrictablesecond{15.54}	& \metrictablesecond{0.520} & \metrictablesecond{0.449} \\
            Difix3D-W (Ours) & 2.4 / 110 & \metrictablefirst{17.64}	& \metrictablefirst{0.577}	& \metrictablefirst{0.414} & \metrictablefirst{17.92} & \metrictablefirst{0.604} & \metrictablefirst{0.419} & \metrictablefirst{18.87} & \metrictablefirst{0.622} &  \metrictablefirst{0.383} & \metrictablefirst{17.54} & \metrictablefirst{0.578} &  \metrictablefirst{0.428} \\
            \bottomrule
		\end{tabular}
	}
    \vspace{-1em}
    \caption{Quantitative results of sparse-view 3D reconstruction methods on the NeRF On-the-go dataset. Performance is ranked by color from \metrictablethird{third} \metrictablesecond{to} \metrictablefirst{first}.}
    \vspace{-1.3em}
	\label{tab:Onthego_quan_results}
\end{table*}
\begin{figure*}[!t]
    \centering
    \includegraphics[width=\linewidth]{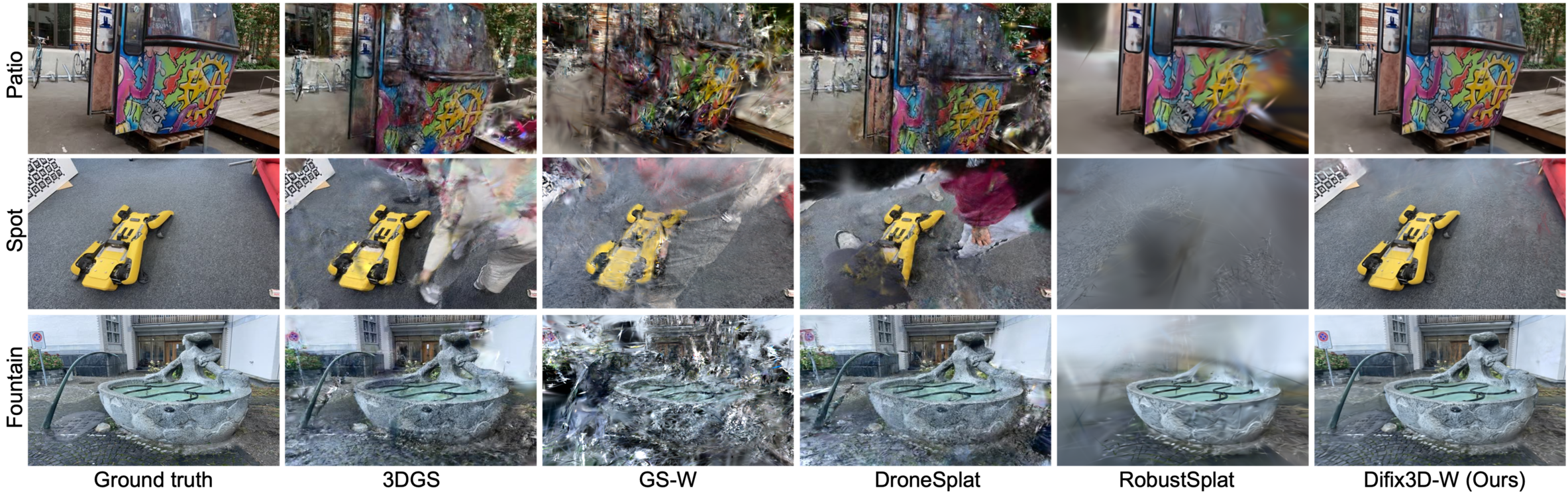}
    \vspace{-2em}
    \caption{Qualitative results on the NeRF On-the-go dataset.} 
    \label{fig:Onthego_qual_results}
    \vspace{-1.3em}
\end{figure*}

\subsection{Reference-Guided Pseudo-Label Synthesis}
\label{subsec:reference-guided_pseudosynthesis}
\noindent \textbf{Pseudo-label synthesis.} Using a sparse set of images presents challenges such as overfitting to the input views and geometric inconsistencies. To mitigate these issues, we generate pseudo-labels using reference views, thereby compensating for insufficient camera viewpoints. Nevertheless, simply using the generated pseudo-label ignores the presence of transient elements. Therefore, we create a refined pseudo-label ($\hat{I}_{\text{pseudo}}$) that references a rendered view from other camera
perspectives along with a transient mask using Eq.~\eqref{eq:diffusion_model}. In
particular, we utilize a mask generator to identify regions that need refinement
from rendered views in order to generate a refined pseudo-label
via Eq.~\eqref{eq:cross-attention}. Lastly, we optimize 3D Gaussians utilizing the
pseudo-label: 
\begin{equation}
  \mathcal{L}_{\text{pseudo}} =  (1-\lambda) \mathcal{L}_1(\hat{I}, \hat{I}_{\text{pseudo}})+\lambda \mathcal{L}_{\text {D-SSIM}}(\hat{I}, \hat{I}_{\text{pseudo}}).
\end{equation} 
Additionally, we employ $\mathcal{L}_{\text{pseudo}}$ to enhance the 3D
representation with $\mathcal{L}_{\text{photo}}$, attaining multi-view
consistency and preventing artifacts and overfitting issues. Note that although
we utilize two vision foundation models in the pipeline, the key distinction of
our work from prior
methods~\cite{liu20243dgs,wu2024reconfusion,kong2025generative},
which also utilize two vision foundation models, is that we generate
a pseudo-label that considers regions that need refinement in scenarios with distractors.

\noindent \textbf{Sparsity-aware Gaussian replication.} 3D point initialization
with COLMAP \cite{schonberger2016structure} from an insufficient set of images
results in sparsity in the Gaussian field. This hinders the rendering quality of
the 3DGS optimization by missing geometric details. Prior
work~\cite{zhu2024fsgs,xu2025ad,zhao2025self} tackles this problem by densifying
the Gaussians via positional gradients with depth supervision. Nonetheless,
simply utilizing positional gradients is insufficient to fix sparsity issues in
the Gaussian field since the positional gradients are entangled with both color
and opacity, 
\begin{equation}
\frac{\partial \mathcal{L}_{\text{photo}}}{\partial\mu} =\frac{\partial \mathcal{L}_{\text{photo}}}{\partial C} \frac{\partial C}{\partial\mu}=\frac{\partial \mathcal{L}_{\text{photo}}}{\partial C}\left(\frac{\partial C}{\partial c} \frac{\partial c}{\partial \mu}+\frac{\partial C}{\partial \alpha} \frac{\partial \alpha}{\partial \mu}\right).
\end{equation} 
Furthermore, as illustrated in Fig.~\ref{fig:sagr}, misaligned Gaussians lead to
artifacts due to redundant Gaussians attempting to blur high-frequency details.
To address this issue, we introduce a strategy that replicates 3D Gaussians by
considering an opacity map to align them. Then, we construct a dense Gaussian
field by filling the sparse region with the new Gaussians. In particular, we use
a accumulated opacity map $\hat{\mathcal{D}}$ as a pixel-wise weight, redefining
$\mathcal{L}_{\text{pseudo}}$ as $\mathcal{L}_{\text{pseudo}} =
\frac{\mathcal{L}_{\text{pseudo}}}{\hat{\mathcal{D}}}$, to adjust the direction
of positional gradients. Finally, by utilizing the accumulated opacity map in the density control process, we solve the sparsity problem     by increasing the density of the Gaussian field.
\begin{table*}[!t]
	\centering
    \footnotesize
    \renewcommand{\arraystretch}{0.9}
    \setlength{\tabcolsep}{10pt}
		\begin{tabular}{cccc c cccc}
			\cmidrule[\heavyrulewidth]{1-4} \cmidrule[\heavyrulewidth]{6-9}
			\multirow{2}{*}{Method} & \multicolumn{3}{c}{Photo Tourism} & \quad &  \multirow{2}{*}{Method} & \multicolumn{3}{c}{LLFF} \\
			&  { PSNR$\uparrow$} & { SSIM$\uparrow$} & { LPIPS$\downarrow$} & & & {PSNR$\uparrow$} & { SSIM$\uparrow$} & { LPIPS$\downarrow$} \\
            \cmidrule[\heavyrulewidth]{1-4}
            \cmidrule[\heavyrulewidth]{6-9}
           NeRF-W \cite{martin2021nerf} & 14.20 & 0.541 & 0.510 & & FreeNeRF \cite{yang2023freenerf}& 19.63 &0.613 & 0.347 \\
           Ha-NeRF \cite{chen2022hallucinated} & 11.73 & 0.483 & 0.381& & SimpleNeRF \cite{somraj2023simplenerf} & 19.24 & \metrictablethird{0.623} & 0.375   \\
           CR-NeRF \cite{yang2023cross} & \metrictablethird{15.08} & \metrictablethird{0.594} & 0.473 & & ZeroNVS \cite{sargent2024zeronvs} & 15.91 &	0.359 &	0.512 \\
            3DGS \cite{kerbl20233d} & 13.99 & 0.456 & 0.499 & & 3DGS \cite{kerbl20233d} & 17.12&	0.467&	0.348\\
           Mip-Splatting \cite{yu2024mip} & 14.13 & 0.461 & 0.481 & & RegNeRF \cite{niemeyer2021regnerf} & 19.08 & 0.587 &	0.374  \\
           DroneSplat \cite{tang2025dronesplat} & 6.288 & 0.232 & 0.630 & & FSGS \cite{zhu2024fsgs} & 19.27 & 0.589 & \metrictablesecond{0.276} \\
           RobustSplat \cite{fu2025robustsplat} & 14.15 & \metrictablesecond{0.644} & \metrictablethird{0.354} & & DiffusionNeRF \cite{wynn2023diffusionerf} & 20.13 & \metrictablesecond{0.631} & 0.344 \\
           WildGaussians \cite{kulhanek2024wildgaussians} & 14.73 & 0.412 & 0.464 & & DropoutGS \cite{park2025dropgaussian} & 18.95& 0.582 &0.335\\
          GS-W \cite{zhang2024gaussian} & 14.03 & 0.482 & 0.467  & & ReconFusion \cite{wu2024reconfusion} & \metrictablethird{21.34} & \metrictablefirst{0.724} & \metrictablefirst{0.203}\\
           SparseGS-W \cite{li2025sparsegs} & \metrictablesecond{19.01} & 0.550 & \metrictablesecond{0.312} & & Difix3D+ \cite{wu2025difix3d+} & \metrictablesecond{22.68}  &	0.571  &	\metrictablethird{0.302}\\
        Difix3D-W (Ours) & \metrictablefirst{19.86} & \metrictablefirst{0.779} & \metrictablefirst{0.306} & & Difix3D-W (Ours) & \metrictablefirst{23.53} & 0.584 & 0.315 \\
            \cmidrule[\heavyrulewidth]{1-4} \cmidrule[\heavyrulewidth]{6-9}
		\end{tabular}
    \vspace{-1.4em}
    \caption{Quantitative results of sparse-view 3D reconstruction methods on the LLFF and Photo Tourism datasets. Performance is ranked by color from \metrictablethird{third} \metrictablesecond{to} \metrictablefirst{first}.}
	\label{tab:photo_llff_quan_results}
    \vspace{-2.3em}
\end{table*}
\subsection{Regularization and Optimization}
\label{subsec:regularization_optimization}
\textbf{Regularization.} We empirically observe that simply using a DM results
in geometric inconsistencies and artifacts. Following recent
methods~\cite{poole2022dreamfusion,shi2023zero123++,sargent2024zeronvs}, we
mitigate these problems by utilizing LoRA~\cite{hu2022lora} in the VAE decoder
to focus on the training scenes. Concretely, we employ LoRA as a test-time
adaption with an SDS loss,
\begin{equation}
  \mathcal{L}_{\mathrm{SDS}}=
  \mathbb{E}_{t \sim \mathcal{U}(0,1),\;
  \epsilon \sim \mathcal{N}(\mathbf{0},\mathbf{I})}
  \left[
  \left\|
  \epsilon_{\theta}(\boldsymbol{x}_t;\hat{I}_{\text{ref}},t)
  -
  \epsilon
  \right\|_2^2
  \right], 
\end{equation}
where $\alpha_t$ and $\epsilon$ are the cumulative product of one minus the
variance schedule and the sample noise, $\boldsymbol{x}_t$ is calculated as $\boldsymbol{x}_t = \alpha_t \hat{I}_p + \sigma_t \epsilon$, and $\hat{I}_p$ denotes the rendered
image in the denoising process. We leverage $\mathcal{L}_{\text{SDS}}$ with
other loss functions to maintain 3D consistency. 

\noindent \textbf{Optimization.} Finally, different from prior approaches, we
optimize the 3D Gaussians and LoRA together, without precomputation, by
utilizing the total loss:
\begin{equation}
  \mathcal{L}_{\text{total}} =
  \mathcal{M}_{t} \odot \mathcal{L}_{\text{GS}}
  +
  \mathcal{L}_{\text{photo}} + \mathcal{L}_{\text{pseudo}} + \mathcal{L}_{\text{SDS}}.
\end{equation} 
During optimization, we periodically replicate and densify the Gaussians to amplify sparse regions of the Gaussian field. Moreover, we include $\mathcal{L}_{\text{pseudo}}$ in the total loss
$\mathcal{L}_{\text{total}}$ after a few iterations to avoid sampling corrupted rendered views from the sparse initialization state.
\begin{table}[!t]
    \centering
    \renewcommand{\arraystretch}{0.3}
    \scriptsize
    \resizebox{0.99\linewidth}{!}{
    \begin{tabular}{clccc}
        \toprule
        & Method & PSNR $\uparrow$ & SSIM $\uparrow$ & LPIPS $\downarrow$ \\ 
        \midrule
        \midrule
        (a) & GS-W~\cite{zhang2024gaussian} & 9.616 & 0.141 & 0.696 \\
        (b) & DroneSplat~\cite{tang2025dronesplat} & 15.75 & 0.544 & 0.436 \\
        (c) & WildGaussians~\cite{kulhanek2024wildgaussians} & 19.14 & 0.623 & 0.357 \\
        (d) & Difix3D-W (Ours) & 23.82 & 0.786 & 0.216 \\
        \bottomrule
    \end{tabular}}
    \vspace{0.5em}
    \includegraphics[width=0.75\linewidth, height=0.1\textheight]{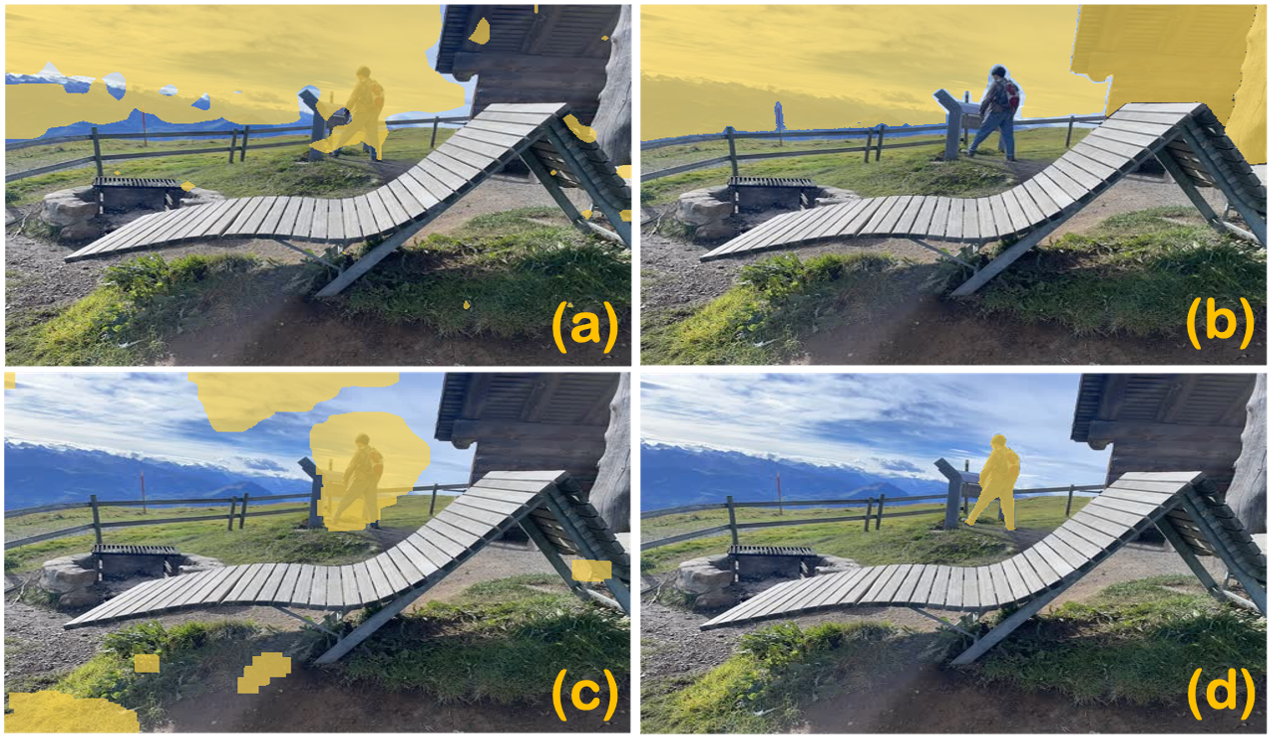}
    \vspace{-1.5em}
    \caption{A comparison of prior methods for generating transient masks on the Mountain scene.}
    \vspace{-2em}
    \label{tab:maskgen}
\end{table}

\begin{table}[!t]
    \centering
    \renewcommand{\arraystretch}{1.1}
    \resizebox{0.99\linewidth}{!}{
    \begin{tabular}{cccccccc}
        \toprule
        \multirow{2}{*}{Method} & \multicolumn{3}{c}{Patio-high} & \multicolumn{3}{c}{Spot}\\
        & PSNR$\uparrow$ & SSIM$\uparrow$ & LPIPS$\downarrow$ & PSNR$\uparrow$ & SSIM$\uparrow$ & LPIPS$\downarrow$\\
        \midrule
        \midrule
        GS-W~\cite{zhang2024gaussian} &  13.47 &	0.335 &	0.470 & 14.33 &	0.534 &	0.480 \\ 
        GS-W~\cite{zhang2024gaussian} + RGVR & 15.31 & 0.428 & 0.419 & 17.04 & 0.613 & 0.433 \\ 
        \midrule
        WildGaussians~\cite{kulhanek2024wildgaussians} &  19.97 & 0.639 & 0.325 & 22.57 & 0.740 & 0.249 \\ 
        WildGaussians~\cite{kulhanek2024wildgaussians} + RGVR &  21.64 & 0.714 & 0.227 & 24.03 & 0.775 & 0.182 \\ 
        \bottomrule
    \end{tabular}}
    \vspace{0.5em}
    \includegraphics[width=1.0\linewidth]{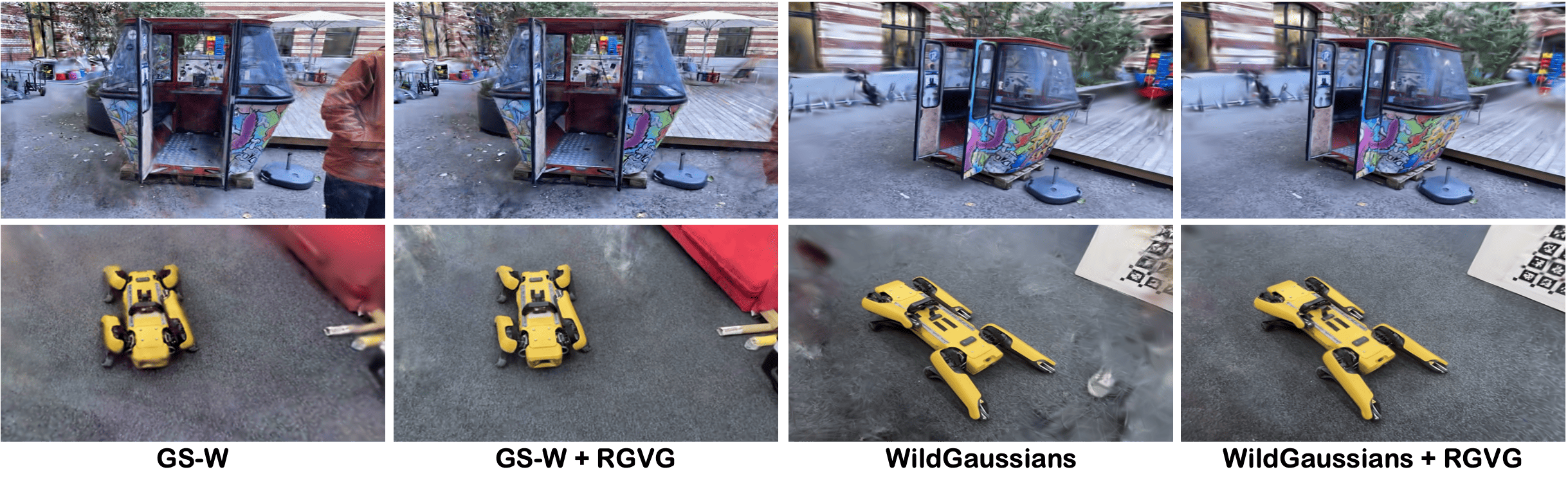}
    \vspace{-3em}
    \caption{An ablation of reference-guided view refinement (RGVR) as a plug-and-play module.}
    \label{tab:plugandplay}
    \vspace{-1.5em}
\end{table}

\section{Experiments}
\noindent \textbf{Datasets, baselines, and metrics.} We evaluate Difix3D-W on
the NeRF On-the-go~\cite{ren2024nerf} and Photo Tourism~\cite{snavely2006photo}
datasets. Following prior
approaches~\cite{kulhanek2024wildgaussians,ren2024nerf,fu2025robustsplat}, we
leverage six scenes with distractors and follow the settings of existing methods
(\eg, image resolution, 3D point initialization, test scenes) on the NeRF
On-the-go dataset. For the Photo Tourism dataset, we utilize three landmark
scenes and abide by the experimental setup of past 
works~\cite{kulhanek2024wildgaussians}. Furthermore, we also evaluate on the
LLFF~\cite{mildenhall2019local} dataset to demonstrate the effectiveness of our
method in constrained scenarios without distractors.  We use COLMAP~\cite{schonberger2016pixelwise,schonberger2016structure} to
initialize the point cloud and camera poses. We compare Difix3D-W with the following past work: (i) fundamental
methods - 3DGS~\cite{kerbl20233d} and Mip-Splatting~\cite{yu2024mip}; (ii)
few-shot-based techniques - RegNeRF~\cite{niemeyer2021regnerf},
DiffusionNeRF~\cite{wynn2023diffusionerf}, FreeNeRF~\cite{yang2023freenerf},
ReconFusion~\cite{wu2024reconfusion}, FSGS~\cite{zhu2024fsgs},
DropoutGS~\cite{park2025dropgaussian}, and Difix3D+~\cite{wu2025difix3d+}; (iii)
in-the-wild-based approaches - NeRF-W~\cite{martin2021nerf},
Ha-NeRF~\cite{chen2022hallucinated}, CR-NeRF~\cite{yang2023cross},
WildGaussian~\cite{kulhanek2024wildgaussians}, GS-W~\cite{zhang2024gaussian},
RobustSplat~\cite{fu2025robustsplat}, DroneSplat~\cite{tang2025dronesplat} and
SparseGS-W~\cite{li2025sparsegs}. Following common practice, we adopt the PSNR,
SSIM~\cite{wang2004image}, and LPIPS~\cite{zhang2018unreasonable} metrics to
assess performance.

\noindent \textbf{Implementation details.} We implemented our method based on
RobustSplat. For the NeRF On-the-go and the LLFF datasets, we employed the
Adam~\cite{kingma2014adam} optimizer without weight decay and set the total
training iterations to 20 K. For the Photo Tourism dataset, we set the total
training iterations to 50 K. We introduced $\mathcal{L}_{\text{pseudo}}$ in the
total loss after 5 K iterations to utilize a reference view. Furthermore, to
generate transient masks, we leveraged the frozen Grounded
SAM~\cite{ren2024grounded}. Although a one-step DM~\cite{wu2025difix3d+} is
redesigned to suit our setting, we followed the default hyperparameter values.
For LoRA, we set the rank to 4 and performed test-time training. Additional
details can be found in the appendix.

\begin{table}[!t]
    \centering
    \renewcommand{\arraystretch}{0.5}
    \resizebox{\linewidth}{!}{
    \begin{tabular}{ccccccc}
        \toprule
        $\hat{I}$ & $\;\;\;\hat{I}_{\text{ref}}$ & $\mathcal{M}_{t}$ & PSNR $\uparrow$ & SSIM $\uparrow$ & LPIPS $\downarrow$ & FID $\downarrow$ \\
        \midrule
        \midrule
        \checkmark & & & 18.89 & 0.597 & 0.304 & 9.130 \\
        \checkmark & \checkmark & & 20.08 & 0.629 & 0.281 & 6.824\\
        \checkmark & \checkmark & \checkmark & 23.03 & 0.732 & 0.269 & 5.295 \\
        \midrule
        \midrule
        \multicolumn{3}{c}{w/o Cross-Attn} & 19.54 & 0.601 & 0.359 & - \\
        \multicolumn{3}{c}{w/ Cross-Attn}  & 23.03 & 0.732 & 0.269 & 5.295 \\
        \bottomrule
    \end{tabular}}
    \vspace{-1em}
    \caption{An analysis of the redesigned one-step DM on the Fountain scene.}
    \label{tab:diffusion}
    \vspace{-1.5em}
\end{table}
\begin{figure}[!t]
    \centering
    \includegraphics[width=\linewidth]{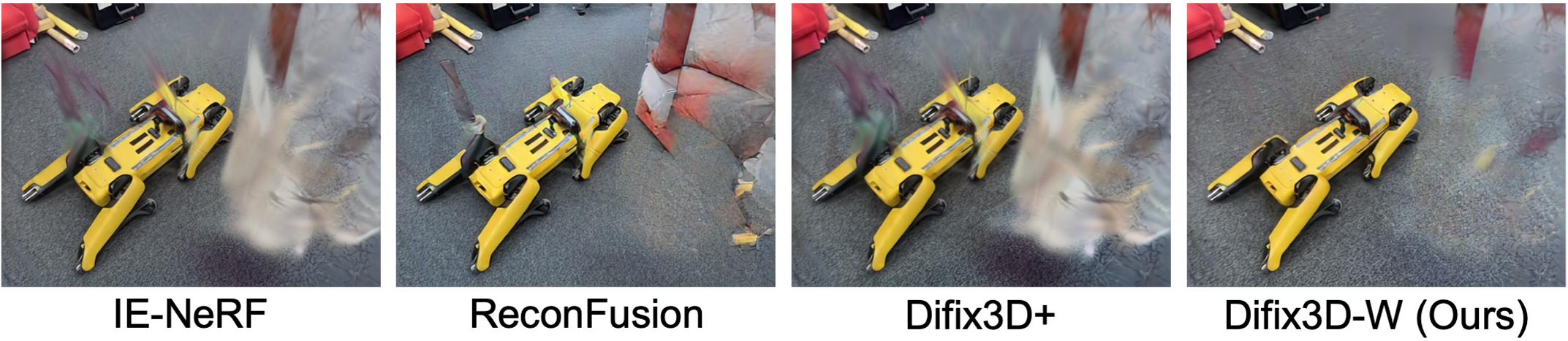}
    \vspace{-2.3em}
    \caption{Comparison of the refined images.}
    \label{fig:refined_images}
    \vspace{-1.35em}
\end{figure}
\begin{figure}[!t]
    \centering
    \includegraphics[width=\linewidth]{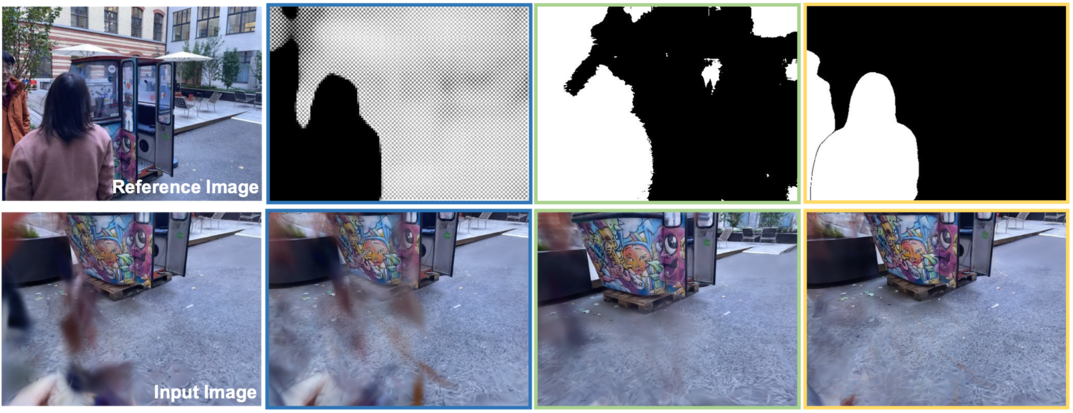}
    \vspace{-2.1em}
    \caption{Robustness of the rendered views.}
    \label{fig:robustness}
    \vspace{-1.5em}
\end{figure}

\noindent \textbf{Comparisons with the state of the art.} We conducted an
extensive evaluation. In Fig.~\ref{fig:Onthego_qual_results}, we observe that
3DGS, GS-W, DroneSplat, and RobustSplat greatly struggle to address distractors. In contrast to prior methods, Difix3D-W achieves high-quality 3D rendering results by distilling insufficient views using a redesigned DM that considers both a reference view and a transient mask. We also achieve competitive performance in
constrained scenes as shown in Tab.~\ref{tab:photo_llff_quan_results}. For the Photo Tourism dataset, CR-NeRF addresses appearance variations by using MLPs to render scenes as continuous functions. Yet, despite their compact
representation, MLPs hinder rendering speed due to the expensive evaluation that
is required for each ray point. Although SparseGS-W also shows fast rendering
speed, it still struggles to address sparsity in a Gaussian field due to its
reliance on the existing positional gradients. However, since Difix3D-W
considers sparse regions to fill the Gaussians and refines a rendered view by
referring to a rendered view from other camera perspectives and a transient
mask, it significantly outperforms the prior state of the art.

\begin{figure}[!t]
    \centering  \includegraphics[width=\linewidth, height=0.15\textheight]{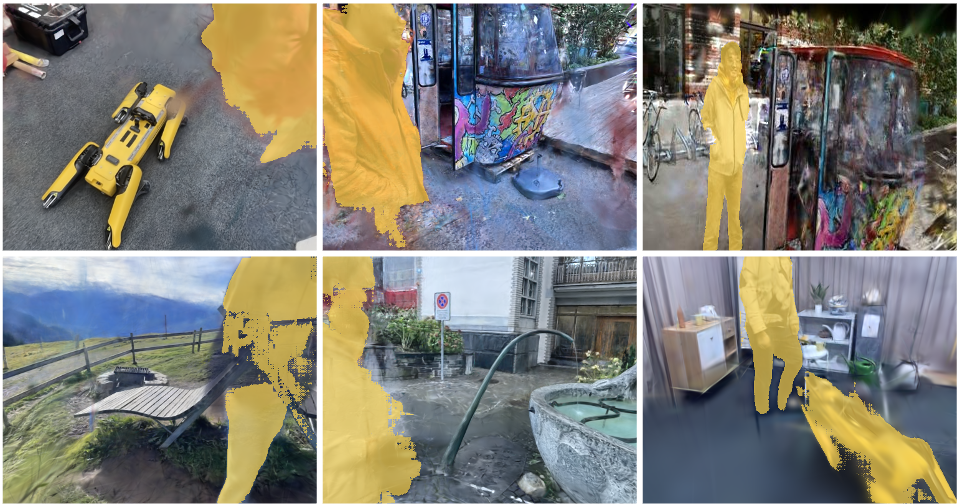}
    \vspace{-2em}
    \captionof{figure}{Examples of transient masks in rendered images on the NeRF On-the-go dataset.}
    \label{fig:mask_corrputed}
    \vspace{-1.2em}
\end{figure}
\begin{table}[!t]
    \centering
    \renewcommand{\arraystretch}{0.7}
    \resizebox{0.99\linewidth}{!}{
    \begin{tabular}{clccc}
        \toprule
        & Method & PSNR $\uparrow$ & SSIM $\uparrow$ & LPIPS $\downarrow$ \\ 
        \midrule
        \midrule
        (a) & w/ Dropout \cite{tang2025dronesplat} & 21.34 & 0.702 & 0.265  \\
        (b) & w/ Unpooling \cite{zhu2024fsgs} & 21.37 & 0.689 & 0.284  \\
        (c) & w/ Sparsity-aware Gaussian replication & 23.82 & 0.786 & 0.216 \\
        \bottomrule
    \end{tabular}}
    \vspace{0.5em}
\includegraphics[width=0.7\linewidth, height=0.1\textheight]{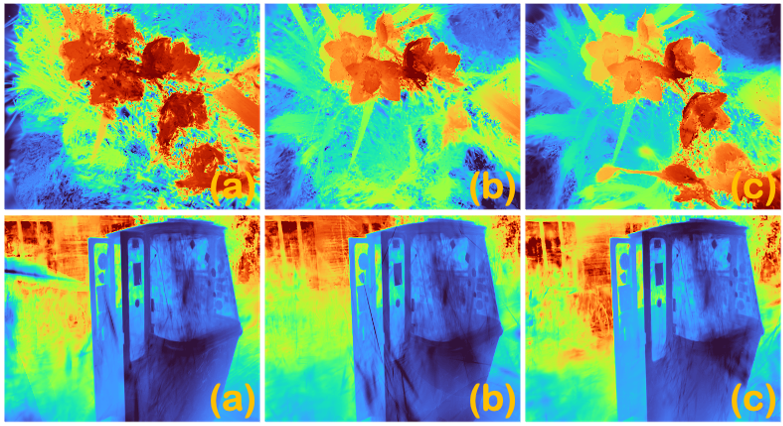}
\vspace{-1.4em}
\caption{An analysis of SAGR on the Orchid and the Patio scenes.}
\label{tab:depth_sagr}
\vspace{-1.5em}
\end{table}
\section{Ablation Study}
\label{sec:ablation_study}
\noindent \textbf{Robustness of view refinement.} We conducted an ablation study to assess the benefit of mask generator approaches,
Tab.~\ref{tab:maskgen}. We observe that the training-based
approach~\cite{zhang2024gaussian} struggles to capture transient elements due to
limited training samples. The heuristic-based technique~\cite{tang2025dronesplat} with
adaptive thresholds tends to excessively mask irrelevant areas, as it lacks
knowledge of transient elements. On the contrary, Difix3D-W captures transient
elements via the power of the vision foundation model, showing generalization
performance. Moreover, our reference-guided view refinement serves as a
plug-and-play module, improving 3D rendering quality as displayed in
Tab.~\ref{tab:plugandplay}.  These results show that reference-guided view
refinement not only efficiently refines rendered views, but it also shows
robustness. Additionally, we analyze our reference-guided view refinement to
assess the capability of the redesigned one-step DM, Tab.~\ref{tab:diffusion}.
We note that incorporating a transient mask to indicate corrupted regions and
refine the rendered view improves 3D rendering quality. Also, compared to prior work \cite{wang2024ie, wu2025difix3d+, wu2024reconfusion}, redesigned one-step DM effectively refine rendered images, as shown in Fig. \ref{fig:refined_images}. Moreover, we observe that although we utilize incorrect transient masks to refine rendered views, redesigned DM robustly show consistent results, as illustrated in Fig. \ref{fig:robustness}.

\begin{table}[!t]
    \centering
    \large
    \renewcommand{\arraystretch}{1.15}
    \resizebox{\linewidth}{!}{
    \begin{tabular}{lccccc}
        \toprule
        & \;\;w/o opacity map \;\;& \;\; w/ opacity map \;\; & PSNR $\uparrow$ & SSIM $\uparrow$ & LPIPS $\downarrow$  \\
        \midrule
        \midrule
         (a) & \checkmark &  &  20.09 & 0.754 & 0.308  \\
         (b) &  &\checkmark  & 22.79 & 0.874 &	0.263  \\
        \bottomrule
    \end{tabular}}
\includegraphics[width=0.8\linewidth]{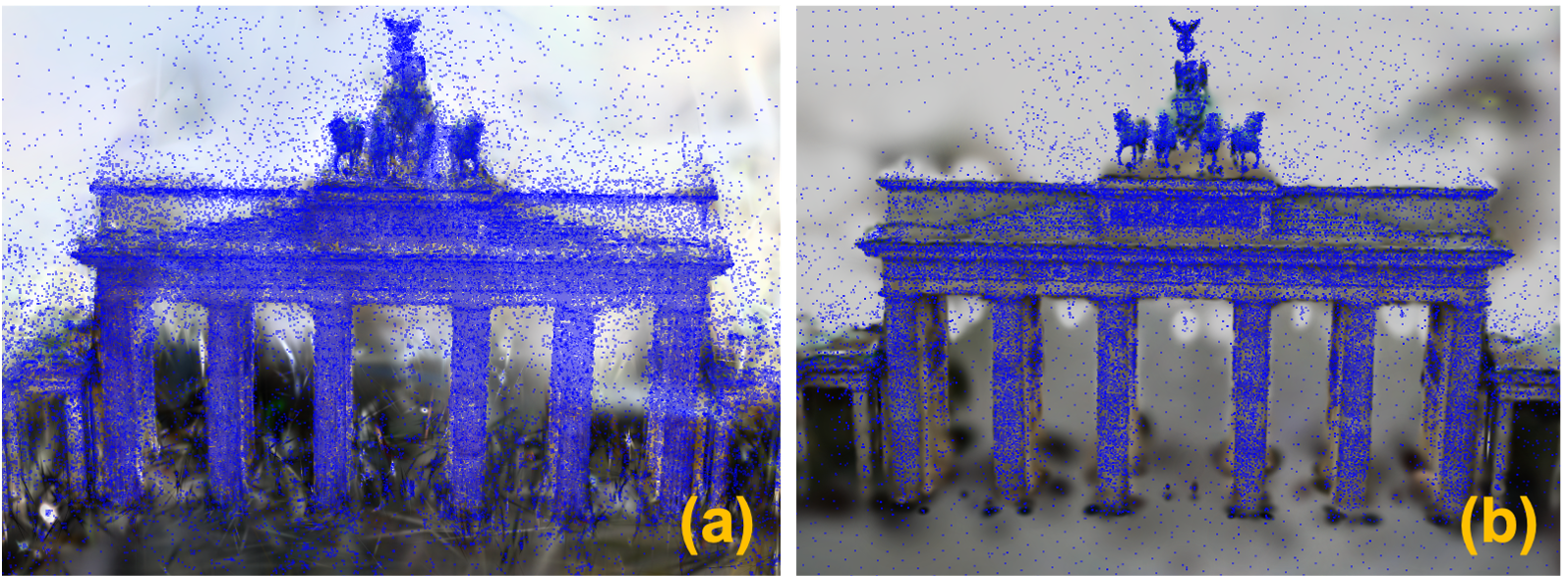}
\vspace{-0.9em}
\caption{An ablation of SAGR. We show the effect of SAGR and visualize the Gaussian anchors.}
\vspace{-1.2em}
\label{tab:anchor}
\end{table}
\begin{table}[!t]
    \centering
  \includegraphics[width=0.8\linewidth]{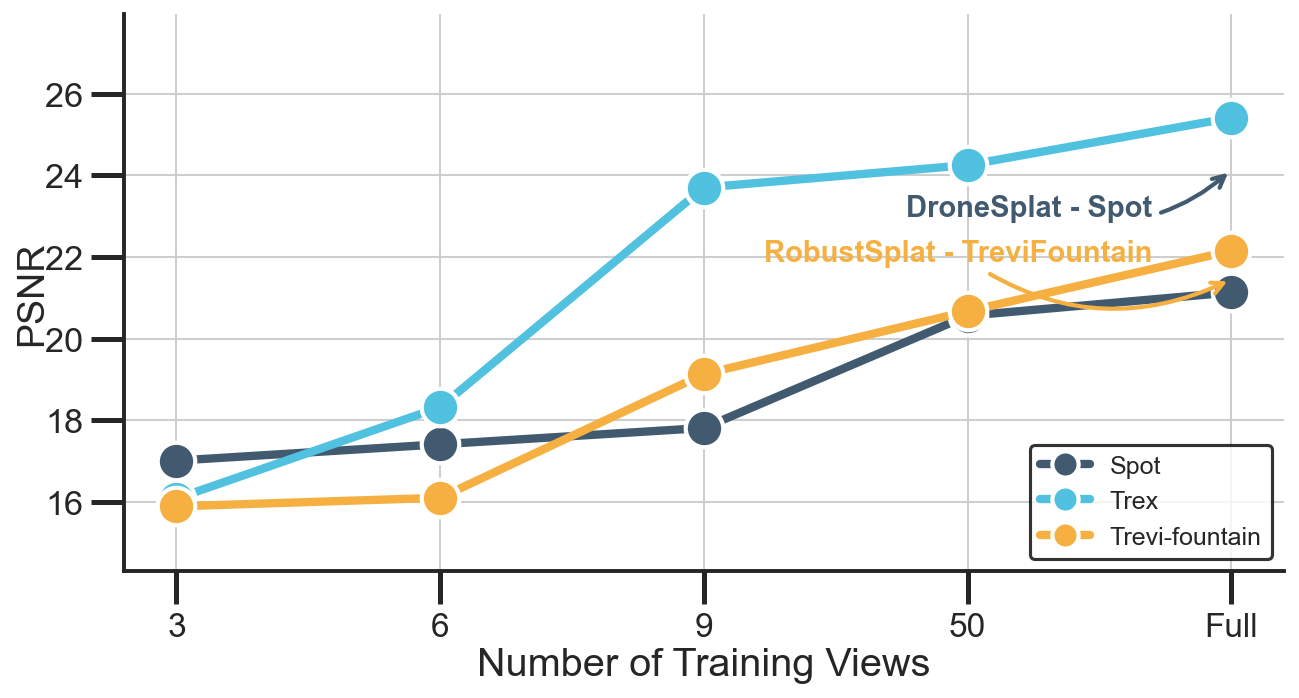}
  \vspace{-1.2em}
  \captionof{figure}{An analysis on the number of training images in the real-world scenarios.}
  \label{fig:scalability}
  \vspace{-1.6em}
\end{table}
\begin{figure}[!t]
\centering
\includegraphics[width=\linewidth]{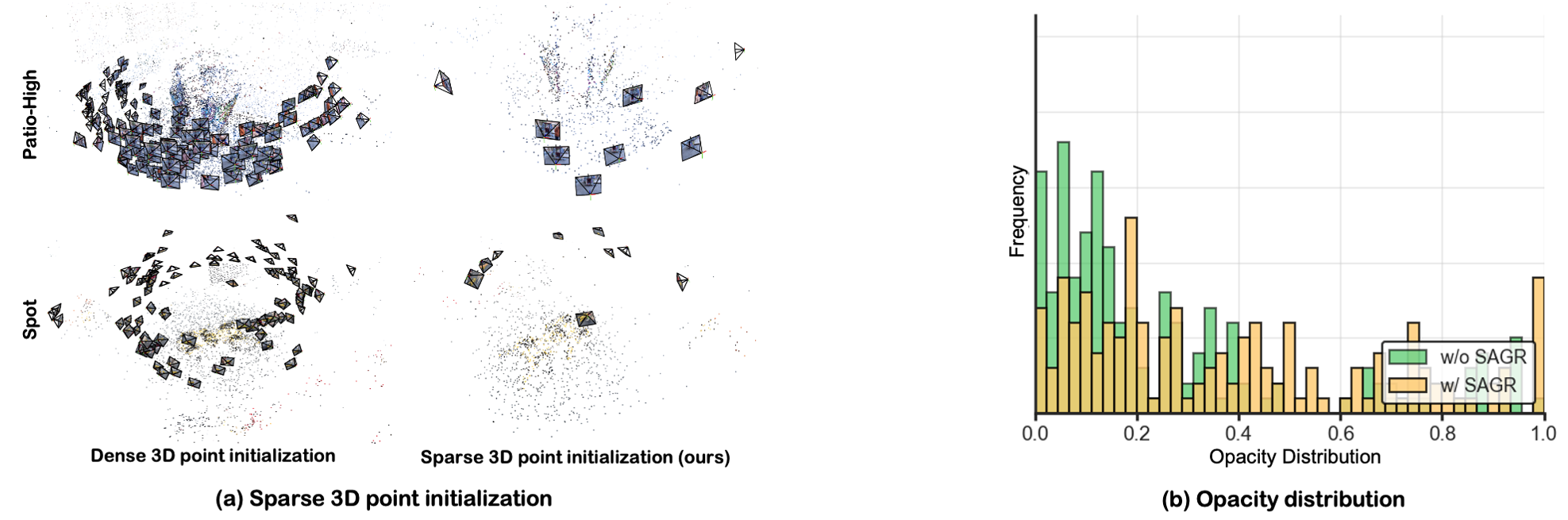}
\vspace{-2em}
\caption{An analysis of 3D Gaussian primitives. (a) The SfM points via COLMAP leveraging 9-views show sparsity compared to the SfM
points resulting from dense image collections. (b) A comparison of the opacity
of distributions w/ and w/o SAGR on the Patio scene.}
\label{fig:point_challenge}
\vspace{-1.5em}
\end{figure}

\noindent \textbf{Effectiveness of pseudo-label synthesis.} We assess the mask
generator for capturing transient elements that are present in the rendered view
as depicted in Fig.~\ref{fig:mask_corrputed}. The results indicate that Grounded
SAM~\cite{ren2024grounded} captures corrupted distractors and shows robustness.
Furthermore, to address the sparsity problem, we utilize an accumulated opacity
map that identifies regions where Gaussians are sparse. Concretely, we conducted
an ablation study on the effect of the sparsity-aware Gaussian replication
(SAGR) strategy in unconstrained real-world scenarios. As reported in
Tab.~\ref{tab:depth_sagr}, compared to existing methods, our SAGR scheme
densifies the Gaussians in the Gaussian field, thus mitigating artifacts and deficient viewpoint issues. Furthermore, we compare our method with
recent density control methods. We observe that
DropoutGS~\cite{park2025dropgaussian} promotes overlapped Gaussians and blurs
rendered views. Although FSGS~\cite{zhu2024fsgs} shows impressive depth maps, it
still struggles to align Gaussians due to the entanglement of the direction of
positional gradients for color and opacity. However, since SAGR considers sparse
regions, we are able to align the Gaussians and maintain 3D consistency,
Tab.~\ref{tab:depth_sagr} and Tab.~\ref{tab:anchor}.  We also study the change
of performance with the number of training views, Fig.~\ref{fig:scalability}. To
highlight the importance of handling sparsity, we visualize the 3D point
initialization in unconstrained conditions,
Fig.~\hyperref[fig:point_challenge]{\ref{fig:point_challenge}-(a)}. The results
demonstrate that employing a lot of training views by referring to diverse
camera perspectives enhances 3D representation. In addition, our results show that SAGR yields Gaussians exhibiting higher opacity values as depicted in Fig.~\hyperref[fig:point_challenge]{\ref{fig:point_challenge}-(b)}. This result indicates that Difix3D-W is able to align the Gaussians. 

\begin{table}[!t]
    \centering
    \renewcommand{\arraystretch}{0.8}
    \resizebox{0.99\linewidth}{!}{
    \begin{tabular}{clccc}
        \toprule
        & Method & PSNR $\uparrow$ & SSIM $\uparrow$ & LPIPS $\downarrow$ \\ 
        \midrule
        \midrule
        (a) & Baseline & 14.33 & 0.412 & 0.514 \\
        (b) & + Reference-guided view refinement & 18.71 & 0.653 & 0.345 \\
        (c) & + Reference-guided pseudo synthesis & 21.85 & 0.661 &	0.297  \\
        (d) & + Sparsity-aware Gaussian replication  & 23.09 & 0.730 &	0.245 \\
        (e) & + $\mathcal{L}_{\text{SDS}}$ & 23.56 & 0.782 &	0.223 \\
        \bottomrule
    \end{tabular}}
\includegraphics[width=0.88\linewidth, height=0.13\textheight]{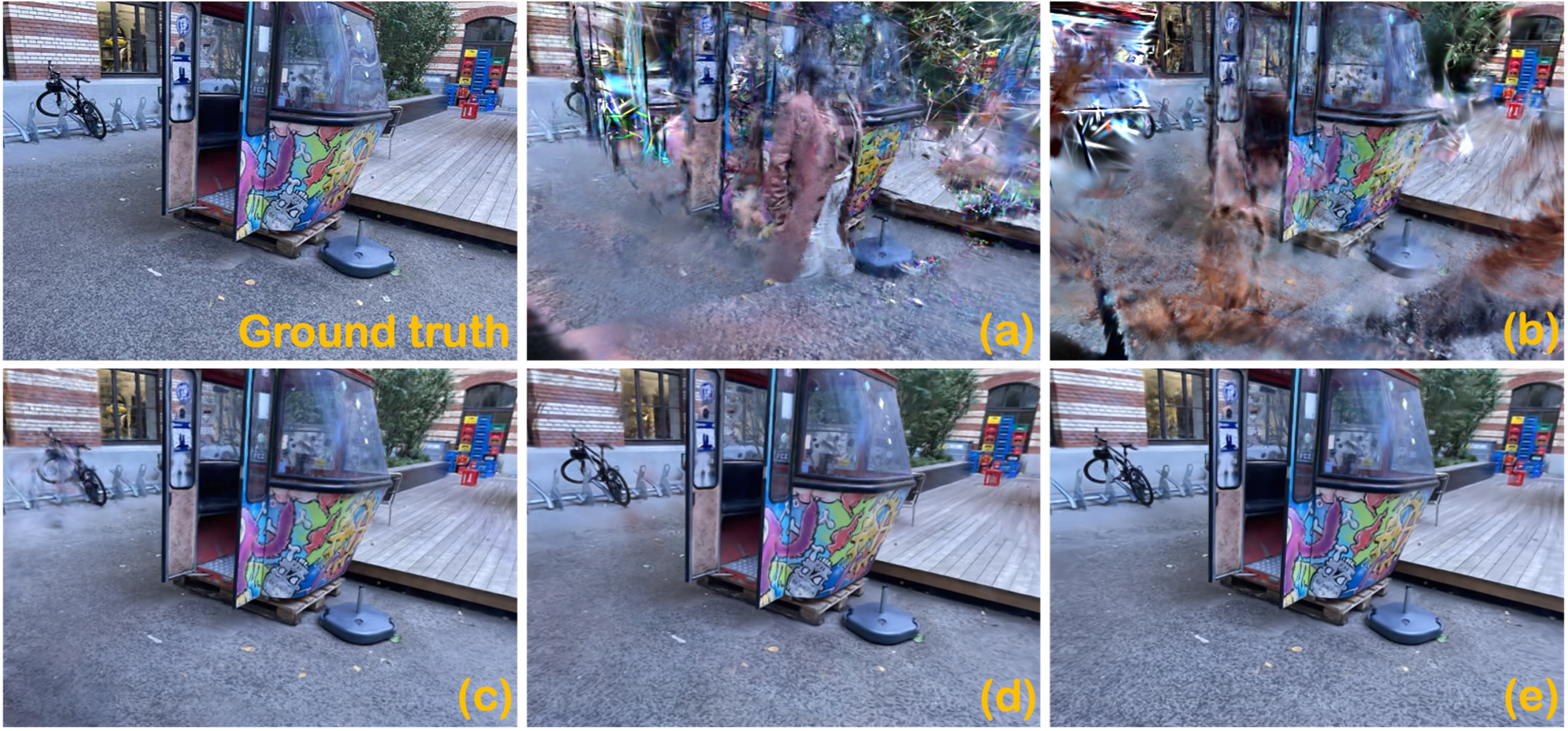}
\vspace{-1em}
\caption{An analysis of each module in Difix3D-W on the Patio-high scene.}
\label{tab:eachmodule}
\vspace{-1.4em}
\end{table}

\noindent \textbf{Analysis of each module.} We evaluated each module
(reference-guided view refinement, reference-guided pseudo-label synthesis,
SAGR, and SDS loss) of Difix3D-W to assess their effectiveness,
Tab.~\ref{tab:eachmodule}. The experimental results indicate that
reference-guided view refinement and reference-guided pseudo-label synthesis
allow us to capture transient elements and refine rendered views using a mask
generator and reference view. Moreover, we observe that adopting LoRA for
test-time training slightly improves the 3D representation in the Gaussian field
as reported in Tab.~\ref{tab:LoRA_Rank}. 

\begin{figure}[!t]
    \centering
    \includegraphics[width=\linewidth]{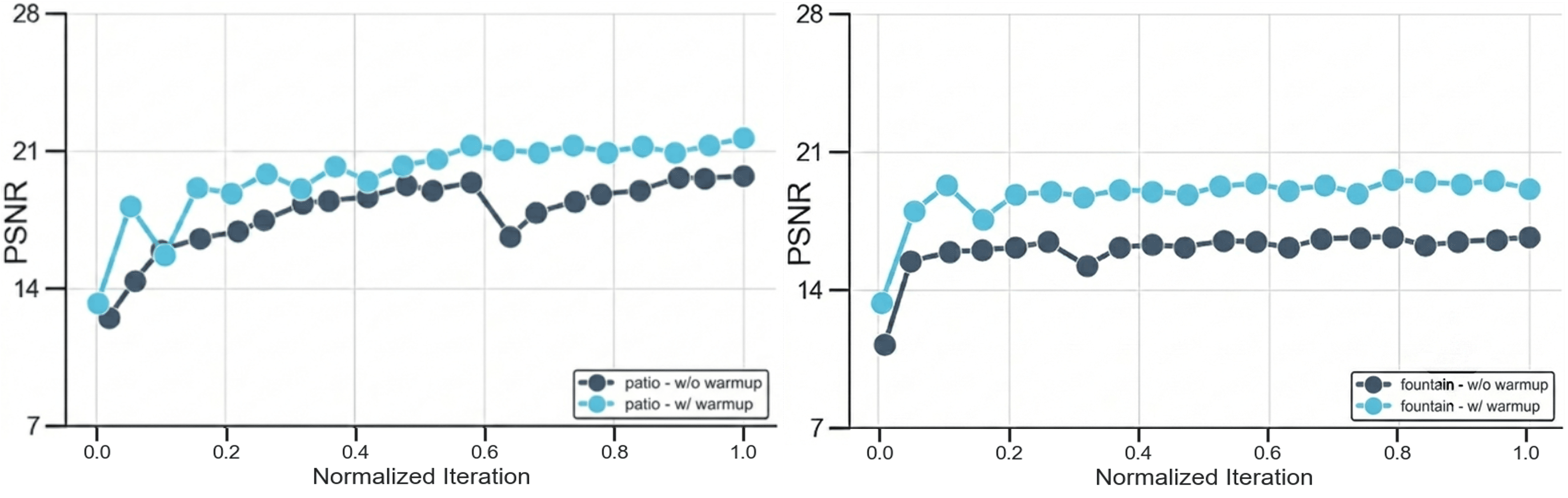}
    \vspace{-2em}
    \caption{A comparison of w/ and w/o a training scheme on
    the Fountain and Patio scene.}
    \label{fig:graph_perform}
    \vspace{-1.5em}
\end{figure}
\begin{table}[!t]
   \centering
    \footnotesize
    \setlength{\tabcolsep}{5pt}
    \renewcommand{\arraystretch}{1}
    \begin{tabular}{cccc}
    \toprule
    Rank (R) & PSNR $\uparrow$ & SSIM $\uparrow$ & LPIPS $\downarrow$  \\
    \midrule
    w/o LoRA \cite{hu2022lora} & 23.08 & 0.579 & 0.317 \\
    R = 4 & 24.32 & 0.605 & 0.293 \\
    R = 8 & 25.21 & 0.605 & 0.287 \\
    R = 16 & 25.37 & 0.601 & 0.294 \\
    \bottomrule
    \end{tabular}
    \vspace{-1em}
    \caption{An analysis of LoRA rank on the Fortress scene.}
    \vspace{-1.8em}
    \label{tab:LoRA_Rank}
\end{table}

We qualitatively and quantitatively observe that incorporating SAGR and
$\mathcal{L}_{\text{SDS}}$ addresses sparsity and maintains 3D consistency in
the Gaussian field. We also investigate whether reference-guided pseudo-label
synthesis should be included at the beginning of training or after a warm-up
phase consisting of a few iterations, Fig.~\ref{fig:graph_perform}. The results
reveal that directly referring to a reference view that is rendered from a
poorly optimized Gaussian field can limit the potential for enhancing 3D
representations. 
\begin{figure}
    \centering    \includegraphics[width=0.7\linewidth]{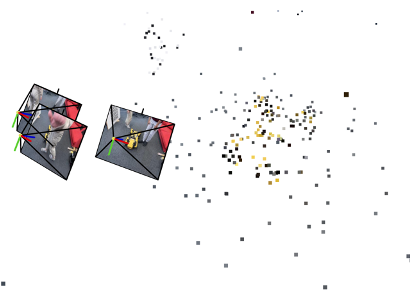}
    \vspace{-1em}
    \caption{Overlapped camera perspectives represent a potential limitation.}
    \vspace{-1.5em}
    \label{fig:limitation}
\end{figure}

\section{Limitations and Conclusion}
\textbf{Limitations.} Although Difix3D-W yields state-of-the-art results on 3D
novel view synthesis for sparse sets of images, there are two limitations: (i)
it can struggle to handle an object that may be static in the whole sequence
(\eg, parked vehicles, standing pedestrians, \etc), which makes it hard to
identify distractors from the background; (ii) as depicted in
Fig.~\ref{fig:limitation}, utilizing overlapped camera perspectives may reduce
the regions available to reference, leading to a degraded 3D representation. To
address these, incorporating prior knowledge about transient elements
along with an additional inpainting model could be a promising direction for
future work.

\noindent \textbf{Conclusion.} In this paper we presented Difix3D-W, a novel
framework for 3D reconstruction from sparse images with distractors.
Difix3D-W consists of a simple and effective reference-guided view refinement that mitigates artifacts in rendered views via a reference view and a transient mask. Furthermore, to tackle sparsity, we introduced reference-guided pseudo-label synthesis, which distills 3D representations using a DM. Our method replicates Gaussians by considering
sparse regions to enhance the 3D representation and construct a dense Gaussian
field. Extensive experimental results on constrained/unconstrained scenarios
show that Difix3D-W achieves significant qualitative and quantitative
improvements over the state of the art. This advancement opens new avenues for
3D novel-view synthesis from a sparse set of real-world images.

\clearpage
\clearpage
\setcounter{page}{1}
\appendix


\maketitle

\section*{Appendix}
In this appendix, we provide an additional discussion, more experimental
results, and other technical details. We organize the appendix into the
following sections.
\begin{itemize}
    \item Sec.~\ref{sub:rel_work}: Related Work
    \item Sec.~\ref{sub:imp_detail}: Implementation Details
    \item Sec.~\ref{sub:exp_detail}: Experiment Results
    \item Sec.~\ref{sub:abl_detail}: Ablation Studies
    \item Sec.~\ref{sub:limit_future}: Limitations and Future Work
    \item Sec.~\ref{sub:use_llm}: Use of Large Language Models
\end{itemize}

\section{Related Work}
\label{sub:rel_work}
\textbf{Sparse-view synthesis for unconstrained scenarios.} Several other
works~\cite{zhang2025rgs,li2025sparsegs} address sparse-view synthesis in
real-world scenarios. However, these methods only focus on appearance variation
settings, not on distractors. Compared to diffusion-based
techniques~\cite{kong2025generative,paliwal2025ri3d,wu2025difix3d+}, Difix3D-W
considers distractors to refine rendered views using a transient mask and a
reference image without penalizing time efficiency.

\noindent \textbf{Feed-forward methods.} Other works generate new Gaussians
utilizing pretrained transformers~\cite{huang2025no,liu2025feed}. Although
these methods show impressive results in constrained scenarios, Difix3D-W
differs as we focus on unconstrained real-world scenes that include distractors.
\section{Implementation Details}
\label{sub:imp_detail}
All optimization and training were performed on a single NVIDIA A40 GPU or RTX 4090 GPU. For our
choice of learning rate, we followed RobustSplat~\cite{fu2025robustsplat}. To
train LoRA, we set the learning rate to 1e-4 across all datasets. For the
DM, we redesigned Difix3D+~\cite{wu2025difix3d+} to utilize a
reference image and a transient mask. Rendered views were refined by randomly
selecting a reference view from diverse camera perspectives. We leveraged
Grounded SAM~\cite{ren2024grounded} to construct transient masks by following
the default settings. To obtain transient masks from Grounded SAM, we used text
descriptions $T_{\text{text}}$ such as humans, robots, and vehicles to capture
distractors across all the datasets. 

For SAGR, we generated new Gaussians by following recent
methods~\cite{fu2025robustsplat, park2026forestsplats} that consider sparse
regions in the 3D Gaussian field. Furthermore, we also employed the original
$\mathcal{L}_\text{photo}$ (2), with a redefined $\mathcal{L}_\text{photo}$, to
train a 3D Gaussian field. To select training views from dense image
collections, we randomly sampled training views by following prior
approaches~\cite{wu2025difix3d+,zhu2024fsgs}. Our approach is shown in
Algorithm~\ref{alg:training_pipeline}.

\noindent \textbf{Architecture details.} We redesigned a one-step DM based on
Difix3D+~\cite{wu2025difix3d+}. We followed the default setting (\eg, guidance,
noise level). However, for cross-attention we utilized a key $\ell_{K}$ and
value $\ell_{V}$ obtained from a reference image. For self-attention, we use the
key $\ell_{K}$ and value $\ell_{V}$ obtained from a rendered image. Furthermore,
we randomly sampled reference images from diverse camera perspectives. Although
the images may be corrupted, Difix3D-W can effectively refine them via its mask
generator.

\noindent \textbf{Baseline details.} To ensure a fair comparison with existing
methods, we reproduced our results using publicly available source code.
Although several
methods~\cite{kong2025generative,paliwal2025ri3d,wu2024reconfusion} are related
to our work, we excluded them if their source code was not available or no
comparable evaluation metrics were reported.

\begin{table}[!t]
    \centering
    \renewcommand{\arraystretch}{0.5}
    \resizebox{\linewidth}{!}{
    \begin{tabular}{cccc}
    \toprule
    Method & PSNR $\uparrow$ & SSIM $\uparrow$ & LPIPS $\downarrow$ \\
    \midrule
    \midrule
    w/o LoRA & 23.08 & 0.579 & 0.317 \\
    w/ LoRA in VAE Decoder & 24.32 & 0.605 & 0.293 \\
    w/ LoRA in U-Net  & 24.35 & 0.607 & 0.291 \\
    \bottomrule
    \end{tabular}}
    \vspace{-1em}
    \caption{An analysis of adopting LoRA on the Fortress scene.}
    \vspace{-1.5em}
    \label{tab:adopt_lora}
\end{table}
\begin{algorithm}[t]
    \SetAlgoLined
    \KwIn{Reference view $\hat{I}_{\text{ref}}$, ground truth $I_{\text{GT}}$, diffusion model $D_{\theta}$, mask generator $S_{\theta}$, camera perspectives $V$, and 3D Gaussian field $R$}
    \KwOut{Rendered view $\hat{I}$, refined view $\tilde{I}_{\text{ref}}$, and refined pseudo-view $\hat{I}_{\text{pseudo}}$}
    \textbf{Initialization:} $ R \gets \text{Structure-from-Motion}$ \cite{schonberger2016structure} \\
    \For {$i$ \textbf{in} ($1,\ldots,N$)}
        {
            $\hat{I} \gets \text{Rasterize}(R, V_i)$ \\
            $\hat{I}_{\text{ref}} \gets \text{Rasterize}(R, V_j)$  \Comment{$i \neq j$} \\
            $\mathcal{M}_t \gets S_{\theta}(\hat{I})$   \\
            $\tilde{I}_{\text{ref}} \gets D_{\theta}(\hat{I}, \hat{I}_{\text{ref}}, \mathcal{M}_t)$ \\
            $\mathcal{L}_{\text{total}} \gets
            \mathcal{M}_{t}\odot \mathcal{L}_{\text{GS}}(\hat{I}, I_{\text{GT}}) +
            \mathcal{L}_{\text{photo}}(\hat{I}, \tilde{I}_{\text{ref}}) + \mathcal{L}_{\text{SDS}}(\hat{I}, \tilde{I}_{\text{ref}})$  \\

            \If{$\text{IsPseudoSynthesis}(i)$ \Comment{Method 4.2} }{
                $\hat{I} \gets \text{Rasterize}(R, V_k)$  \Comment{$i \neq j \neq k$} \\
                $\hat{I}_{\text{pseudo}} \gets D_{\theta}(\hat{I}, \hat{I}_{\text{ref}}, \mathcal{M}_t)$ \\
                $\mathcal{L}_{\text{total}} \gets 
                \mathcal{M}_{t}\odot \mathcal{L}_{\text{GS}}(\hat{I}, I_{\text{GT}}) +\mathcal{L}_{\text{photo}}(\hat{I}, \tilde{I}_{\text{ref}}) + \mathcal{L}_{\text{pseudo}}(\hat{I}, \hat{I}_{\text{pseudo}}) + \mathcal{L}_{\text{SDS}}(\hat{I}, \tilde{I}_{\text{ref}})$\\

                \If{$\text{IsReplication}(i)$ }{
                    $\hat{D} \gets \text{Rasterize}(R, V_k)$ \\
                    $\text{Densification}(R)$  \\
                }
            }
        }
    \caption{Training Pipeline}
    \label{alg:training_pipeline}
\end{algorithm}

\section{Experiment Results}
\label{sub:exp_detail}
\textbf{Quantitative results.} Additional quantitative experiment results are
shown in Tab.~\ref{tab:onthego_3view}, Tab.~\ref{tab:onthego_6view}, and
Tab.~\ref{tab:onthego_9view}. Difix3D-W consistently provides impressive
results compared to existing methods in real-world scenarios that include
distractors. Although Difix3D+~\cite{wu2025difix3d+} also shows high-quality
results, it struggles to refine rendered views in real-world scenarios due to a
lack of knowledge of distractors. Moreover, we observed that leveraging
semantic-level masking better captures transient elements compared to
GS-W~\cite{zhang2024gaussian} and DroneSplat~\cite{tang2025dronesplat}. In
particular, those frameworks have difficulties in capturing transient elements
that have similar colors to the static background.

\begin{figure*}[!t]
    \centering
    \includegraphics[width=\textwidth]{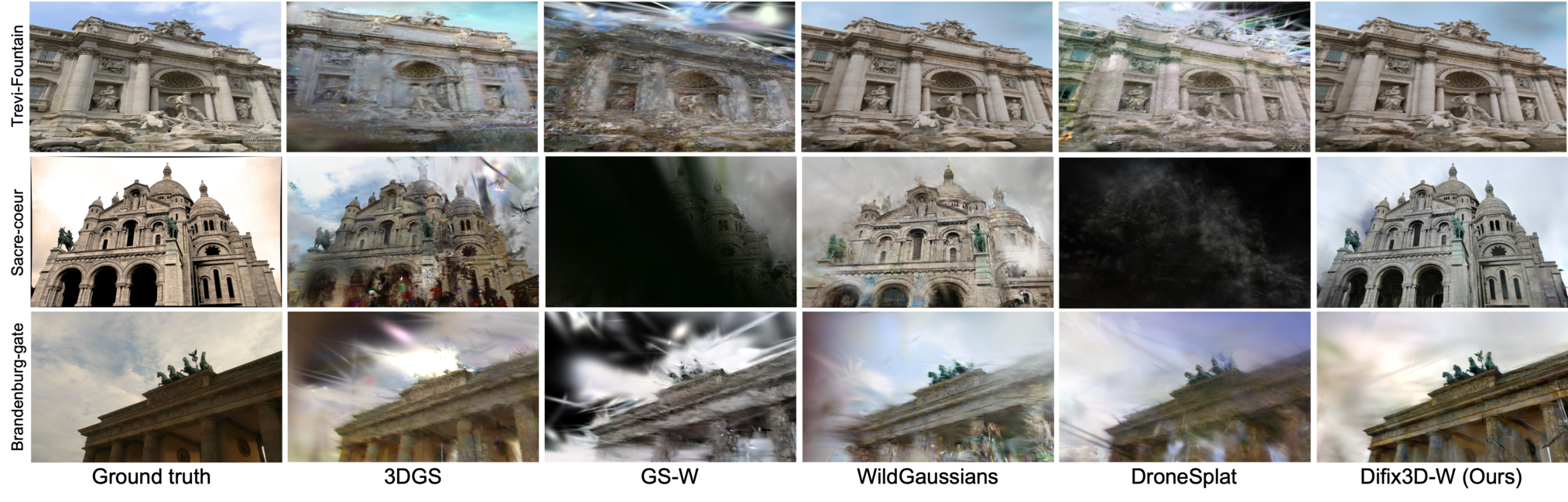}
    \caption{Qualitative results on the Photo Tourism dataset.}
    \label{fig:photo_results}
\end{figure*}

\begin{figure*}[!t]
    \centering
    \includegraphics[width=\textwidth]{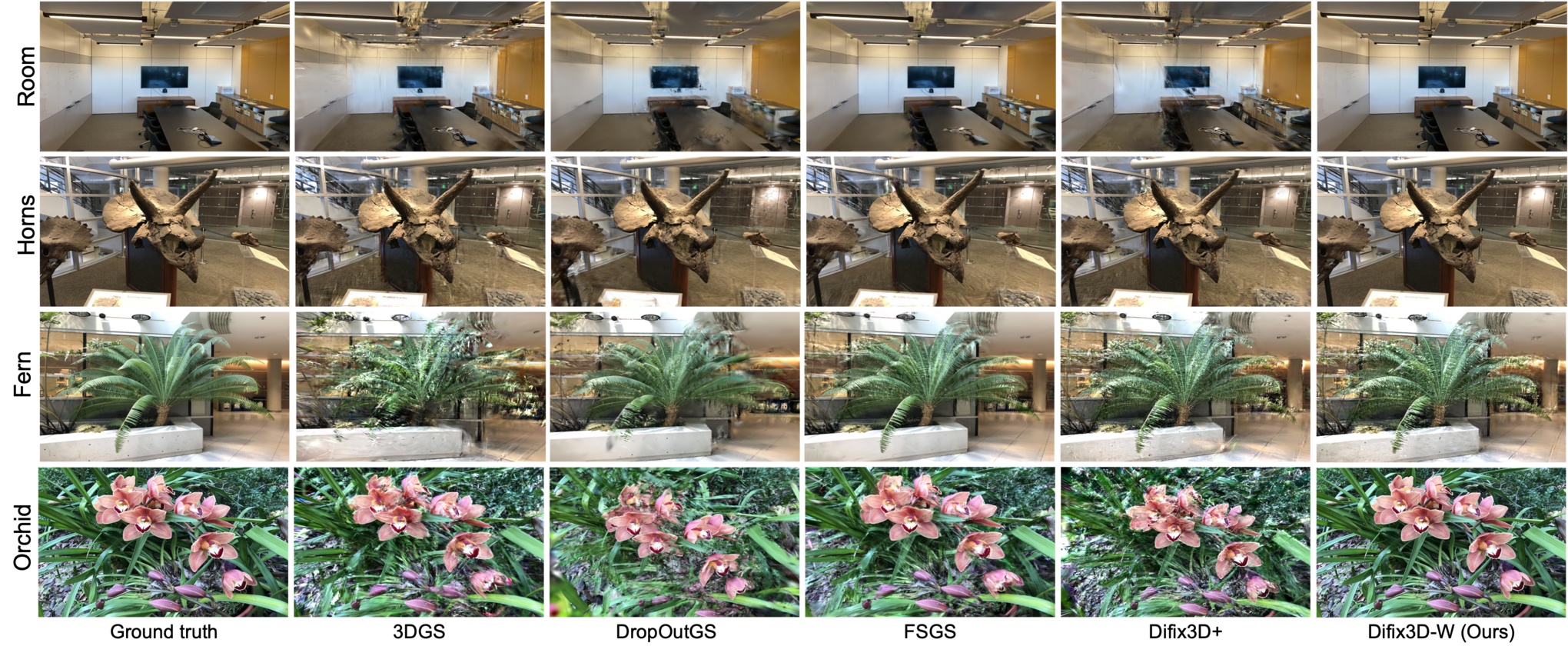}
    \caption{Qualitative results on the LLFF dataset.} 
    \label{fig:llff_results}
\end{figure*}

\noindent \textbf{Qualitative results.} We provide more detailed qualitative
experiment results in Fig.~\ref{fig:photo_results} and
Fig.~\ref{fig:llff_results}. We observe that although Difix3D-W does not
utilize appearance embeddings, in comparison to other techniques it demonstrates
robustness on the Photo Tourism dataset. Moreover, our method shows competitive
results by capturing more details compared to other methods. The results
demonstrate that Difix3D-W can be utilized in unconstrained or constrained
real-world scenarios.

\vspace{-1em}
\begin{table}[!t]
    \centering
    \setlength{\tabcolsep}{15pt}
    \renewcommand{\arraystretch}{0.75}
    \resizebox{\linewidth}{!}{
    \begin{tabular}{cccccc}
    \toprule
    Scene  & Method &  Memory  & Scene &  Method & Memory\\ 
    \midrule 
    \midrule
    \multirow{3}{*}{{\shortstack{Patio}}} & WildGaussians ~\cite{kulhanek2024wildgaussians}  &  57.31 & \multirow{3}{*}{Mountain} & WildGaussians~\cite{kulhanek2024wildgaussians}  & 67.86 \\
        & Difix3D+~\cite{wu2025difix3d+} &  201.93 & & Difix3D+~\cite{wu2025difix3d+}   & 236.51  \\
        & Difix3D-W (Ours)  & 189.17 & & Difix3D-W (Ours) & 209.26 \\
        \midrule 
        \midrule
        \multirow{3}{*}{\shortstack{Corner}} &WildGaussians~\cite{kulhanek2024wildgaussians} & 57.87 & \multirow{3}{*}{\shortstack{Spot}} & WildGaussians~\cite{kulhanek2024wildgaussians} & 14.83 \\
        & Difix3D+~\cite{wu2025difix3d+} & 171.26 & & Difix3D+~\cite{wu2025difix3d+} &  182.67 \\
        &  Difix3D-W (Ours) & 153.48 & & Difix3D-W (Ours) & 143.81 \\
    \bottomrule
    \end{tabular}}
    \vspace{-1em}
    \caption{An analysis of memory usage on the NeRF On-the-go dataset.}
    \vspace{-1.2em}
    \label{tab:onthego_memory}
\end{table}
\begin{figure}[!t]
\centering
\includegraphics[width=\linewidth]{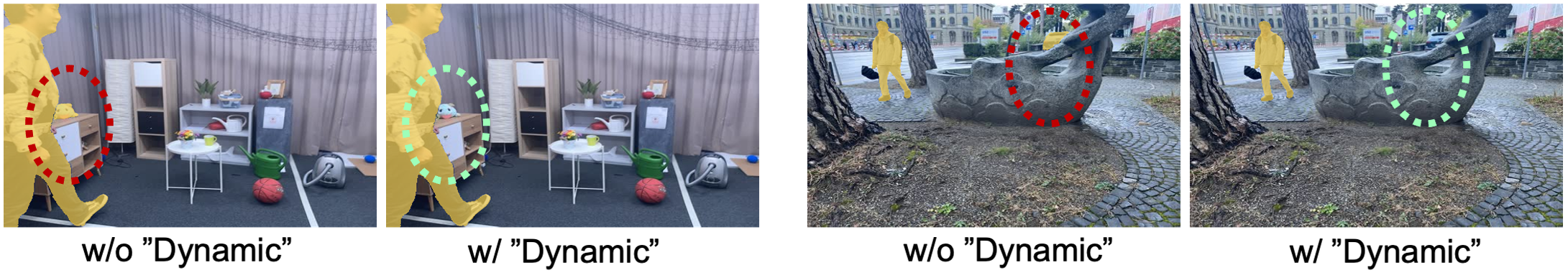}
\vspace{-2em}
\caption{An ablation of robustness on text descriptions.}
\vspace{-1em}
\label{fig:text_descriptions}
\end{figure}

\section{Ablation Studies}
\label{sub:abl_detail}
\textbf{Adopting LoRA.} We examined whether incorporating LoRA in the DM is
beneficial. The results indicate that employing LoRA in the U-Net slightly
improved rendering results as depicted in Tab.~\ref{tab:adopt_lora}. Even though
the performance gain is marginal, we applied LoRA to the VAE decoder, which also
adapts well to the training scene.

\noindent \textbf{Memory efficiency.} We reported memory usage with 9-view
training on the NeRF On-the-go dataset, Tab.~\ref{tab:onthego_memory}. Although
Difix3D-W utilizes more memory than other
methods~\cite{kulhanek2024wildgaussians,wu2025difix3d+}, these approaches
struggle to render high-quality results. Nonetheless, Difix3D-W effectively
aligns Gaussians in the Gaussian field, showing impressive results. Moreover,
the results indicate that SAGR spreads the Gaussians uniformly well in the
Gaussian field and avoids overfitting.

\noindent \textbf{Robustness for text descriptions.} We examined whether the 
mask generator is robust if the text description is changed,
Fig.~\ref{fig:text_descriptions}. We observe that including the word ``dynamic''
in $T_{\text{text}}$ effectively captured transient elements. Inspired by this,
we included the word ``dynamic'' in the text descriptions.

\begin{figure}[!t]
\centering
\includegraphics[width=\linewidth]{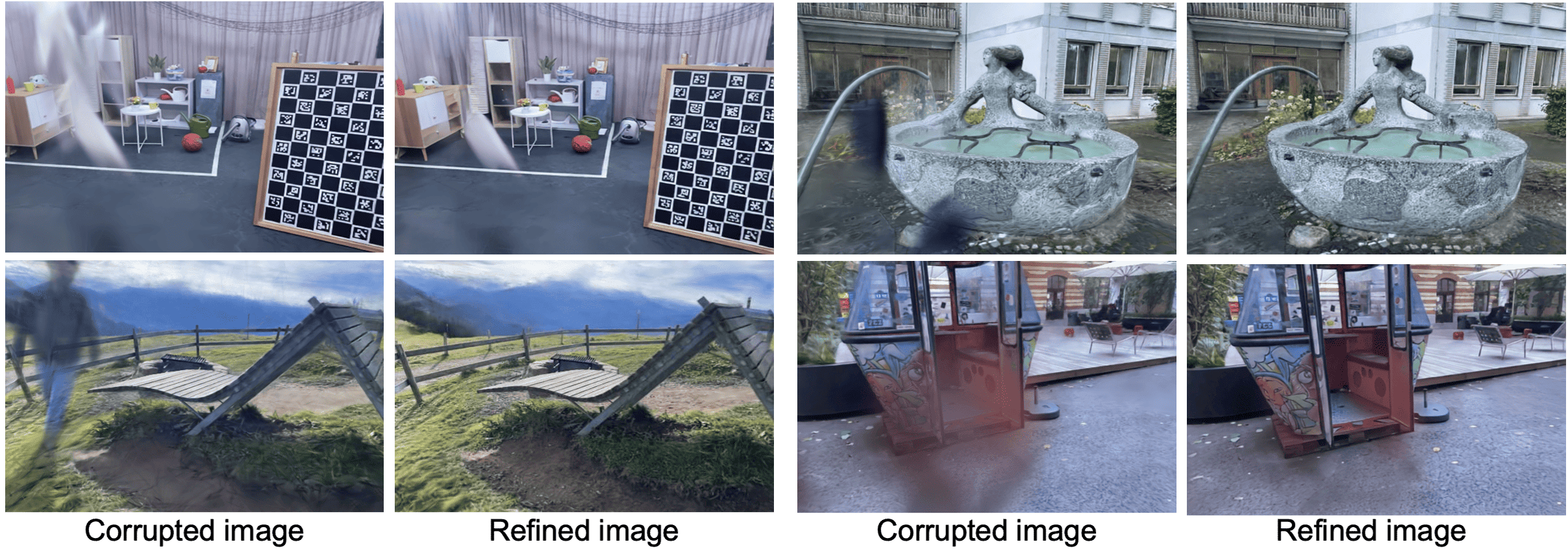}
\caption{Qualitative results showing the refined images from the corrupted
images.}
\vspace{-1.5em}
\label{fig:corrupt_to_refine}
\end{figure}

\noindent \textbf{Refined views.} To qualitatively evaluate the DM, we
visualized refined views from corrupted images using the redesigned DM. As
displayed Fig.~\ref{fig:corrupt_to_refine}, the results show that Difix3D-W can
effectively improve corrupted images and distill a 3D representation in the
Gaussian field by using the DM.

\section{Limitations and Future Work}
\label{sub:limit_future}
\textbf{Limitations.} Although Difix3D-W shows impressive results, it has two
limitations. First, despite the DM refining corrupted images, the refined images
may contain a slight amount of noise as highlighted in
Fig.~\ref{fig:corrupt_to_refine}. Second, although the mask generator is good at
capturing noise and blur in the rendered images, it can miss other artifacts. To
address this problem, a super-resolution model can be used to refine corrupted
regions.

\vspace{0.2\baselineskip}
\noindent \textbf{Future Work.} Difix3D-W provides high-quality rendering
results in constrained and unconstrained real-world scenarios. Nonetheless,
since our framework utilizes a Gaussian field, it may struggle to bridge the gap
between physical AI and real-world scenarios due to the presence of artifacts.
We hope to close this gap by exploring mesh-based techniques in the future.
\section{Use of Large Language Models}
\label{sub:use_llm}
We employed a large language model for copy editing, including grammar checking,
wording refinement, and minor improvements in style and clarity. This was done
after we had completed the scientific content, methodology, analyses, and
conclusions.
\clearpage
\clearpage
\begin{table*}[!t]
    \centering
    \setlength{\tabcolsep}{0.8pt}
    \renewcommand{\arraystretch}{1}
    \resizebox{\linewidth}{!}{\begin{tabular}{cccccccccccccccccccccccccccc}
    \toprule
    \multirow{2}{*}{Method} 
    & \multicolumn{3}{c}{Mountain} 
    & \multicolumn{3}{c}{Fountain} 
    & \multicolumn{3}{c}{Corner} 
    & \multicolumn{3}{c}{Patio} 
    & \multicolumn{3}{c}{Spot} 
    & \multicolumn{3}{c}{Patio-High}\\
    & PSNR $\uparrow$ & SSIM $\uparrow$ & LPIPS $\downarrow$ 
    & PSNR $\uparrow$ & SSIM $\uparrow$ & LPIPS $\downarrow$ 
    & PSNR $\uparrow$ & SSIM $\uparrow$ & LPIPS $\downarrow$
    & PSNR $\uparrow$ & SSIM $\uparrow$ & LPIPS $\downarrow$ 
    & PSNR $\uparrow$ & SSIM $\uparrow$ & LPIPS $\downarrow$ 
    & PSNR $\uparrow$ & SSIM $\uparrow$ & LPIPS $\downarrow$ \\
    \midrule
    \midrule
    3DGS \cite{kerbl20233d}  
    & 9.323 & 0.144 & 0.599 & 9.231 & 0.155 & 0.573 & 13.98 & 0.373 & \metrictablethird{0.462} & 14.01 & 0.410 & 0.452 & 12.76 & 0.375 & 0.693 & 10.74 & 0.179 & 0.594 \\
    RobustSplats \cite{fu2025robustsplat}  
    & 6.008 & 0.060 & 0.776 & 11.82 & 0.335 & 0.605 & 13.40 & 0.350 & 0.540 & 12.76 & 0.375 & 0.603 & \metrictablethird{16.08} & \metrictablethird{0.427} & 0.671 & \metrictablethird{12.76} & 0.283 & 0.575 \\
    GS-W \cite{zhang2024gaussian}  
    & 11.23 & 0.279 & 0.599 & 6.852 & 0.076 & 0.792 & 12.56 & 0.417 & 0.583 & 10.50 & 0.241 & 0.591 & 14.02 & 0.417 & \metrictablethird{0.664} & 11.83 & \metrictablethird{0.351} & \metrictablethird{0.536} \\
    DroneSplat \cite{tang2025dronesplat}  
    & 11.34 & 0.234 & 0.515 & 9.524 & 0.168 & 0.583 & 15.19 & \metrictablethird{0.422} & 0.398 & 9.305 & 0.131 & 0.594 & 12.83 & 0.325 & \metrictablesecond{0.662} & 11.66 & 0.219 & 0.583 \\
    WildGaussians \cite{kulhanek2024wildgaussians}  
    & \metrictablethird{11.75} & \metrictablethird{0.545} & \metrictablethird{0.396} & \metrictablethird{12.46} & \metrictablethird{0.359} & \metrictablethird{0.530} & \metrictablethird{13.68} & 0.408 & 0.497 & \metrictablethird{14.62} & \metrictablethird{0.521} & \metrictablethird{0.338} & 15.40 & 0.419 & 0.674 & 12.71 & 0.289 & \metrictablesecond{0.564} \\
    Difix3D+ \cite{wu2025difix3d+}  
    & \metrictablesecond{15.21} & \metrictablesecond{0.608} & \metrictablesecond{0.329} & \metrictablesecond{14.71} & \metrictablesecond{0.546} & \metrictablesecond{0.399} & \metrictablesecond{17.13} & \metrictablesecond{0.510} & \metrictablesecond{0.416} & \metrictablesecond{17.03} & \metrictablesecond{0.497} & \metrictablesecond{0.445} & \metrictablesecond{16.37} & \metrictablesecond{0.439} & 0.680 & \metrictablesecond{14.53} & \metrictablesecond{0.452} & 0.621 \\
    Difix3D-W (Ours)  
    & \metrictablefirst{19.37} & \metrictablefirst{0.685} & \metrictablefirst{0.310} & \metrictablefirst{16.12} & \metrictablefirst{0.581} & \metrictablefirst{0.310} & \metrictablefirst{19.65} & \metrictablefirst{0.631} & \metrictablefirst{0.257} & \metrictablefirst{17.89} & \metrictablefirst{0.610} & \metrictablefirst{0.431} & \metrictablefirst{17.01} & \metrictablefirst{0.472} & \metrictablefirst{0.646} & \metrictablefirst{15.82} & \metrictablefirst{0.488} & \metrictablefirst{0.534} \\
    \bottomrule
    \end{tabular}}
    \caption{Quantitative results on the NeRF On-the-go dataset under a 3-view training setting. Performance is ranked by color from \metrictablethird{third} \metrictablesecond{to} \metrictablefirst{first}.}
    \label{tab:onthego_3view}
\end{table*}
\begin{table*}[!t]
    \centering
    \setlength{\tabcolsep}{0.8pt}
    \renewcommand{\arraystretch}{1}
    \resizebox{\linewidth}{!}{\begin{tabular}{cccccccccccccccccccccccccccc}
    \toprule
    \multirow{2}{*}{Method} 
    & \multicolumn{3}{c}{Mountain} 
    & \multicolumn{3}{c}{Fountain} 
    & \multicolumn{3}{c}{Corner} 
    & \multicolumn{3}{c}{Patio} 
    & \multicolumn{3}{c}{Spot} 
    & \multicolumn{3}{c}{Patio-High}\\
    & PSNR $\uparrow$ & SSIM $\uparrow$ & LPIPS $\downarrow$ 
    & PSNR $\uparrow$ & SSIM $\uparrow$ & LPIPS $\downarrow$ 
    & PSNR $\uparrow$ & SSIM $\uparrow$ & LPIPS $\downarrow$
    & PSNR $\uparrow$ & SSIM $\uparrow$ & LPIPS $\downarrow$ 
    & PSNR $\uparrow$ & SSIM $\uparrow$ & LPIPS $\downarrow$ 
    & PSNR $\uparrow$ & SSIM $\uparrow$ & LPIPS $\downarrow$ \\
    \midrule
    \midrule
    3DGS \cite{kerbl20233d}  
    & \metrictablesecond{14.52} & 0.332 & 0.398 & 9.281 & 0.194 & 0.596 & 15.16 & 0.420 & 0.404 & 12.98 & 0.341 & 0.483 & 14.26 & 0.307 & 0.678 & 11.21 & 0.263 & 0.571 \\
    RobustSplats \cite{fu2025robustsplat}  
    & 10.74 & 0.309 & 0.733 & 12.30 & 0.346 & 0.593 & 13.69 & \metrictablethird{0.491} & 0.513 & 15.18 & 0.570 & 0.426 & \metrictablethird{15.99} & 0.424 & \metrictablethird{0.671} & 14.33 & 0.413 & 0.586 \\
    GS-W \cite{zhang2024gaussian}  
    & 12.96 & 0.351 & 0.519 & 11.17 & 0.177 & 0.617 & 14.33 & 0.480 & 0.535 & 13.47 & 0.335 & 0.471 & 15.61 & \metrictablesecond{0.469} & 0.699 & 14.25 & 0.410 & 0.556 \\
    DroneSplat \cite{tang2025dronesplat}  
    & 12.96 & 0.316 & 0.481 & 11.94 & 0.271 & 0.478 & \metrictablethird{15.58} & 0.489 & \metrictablethird{0.361} & 13.09 & 0.357 & \metrictablethird{0.401} & 14.78 & 0.398 & 0.698 & 12.10 & 0.276 & \metrictablefirst{0.531} \\
    WildGaussians \cite{kulhanek2024wildgaussians} 
    & 12.17 & \metrictablethird{0.553} & \metrictablethird{0.393} & \metrictablethird{13.12} & \metrictablethird{0.390} & \metrictablethird{0.532} & 14.48 & 0.483 & 0.411 & \metrictablethird{15.93} & \metrictablethird{0.595} & \metrictablefirst{0.399} & 15.42 & \metrictablethird{0.460} & 0.689 & \metrictablethird{14.41} & \metrictablethird{0.423} & \metrictablesecond{0.540} \\
    Difix3D+ \cite{wu2025difix3d+}  
    & \metrictablethird{14.06} & \metrictablesecond{0.595} & \metrictablesecond{0.334} & \metrictablesecond{15.82} & \metrictablesecond{0.571} & \metrictablesecond{0.375} & \metrictablesecond{18.04} & \metrictablesecond{0.597} & \metrictablethird{0.290} & \metrictablesecond{17.24} & \metrictablesecond{0.611} & 0.486 & \metrictablesecond{16.96} & \metrictablesecond{0.469} & \metrictablesecond{0.672} & \metrictablesecond{15.94} & \metrictablesecond{0.508} & 0.607 \\
    Difix3D-W (Ours)  
    & \metrictablefirst{19.63} & \metrictablefirst{0.694} & \metrictablefirst{0.308} & \metrictablefirst{16.38} & \metrictablefirst{0.602} & \metrictablefirst{0.307} & \metrictablefirst{19.68} & \metrictablefirst{0.640} & \metrictablefirst{0.256} & \metrictablefirst{18.10} & \metrictablefirst{0.688} & \metrictablethird{0.429} & \metrictablefirst{17.41} & \metrictablefirst{0.490} & \metrictablefirst{0.661} & \metrictablefirst{16.37} & \metrictablefirst{0.512} & \metrictablethird{0.554} \\
    \bottomrule
    \end{tabular}}
    \caption{Quantitative results on the NeRF On-the-go dataset under a 6-view training setting. Performance is ranked by color from \metrictablethird{third} \metrictablesecond{to} \metrictablefirst{first}.}
    \label{tab:onthego_6view}
\end{table*}
\begin{table*}[!t]
    \centering
    \setlength{\tabcolsep}{0.8pt}
    \renewcommand{\arraystretch}{1}
    \resizebox{\linewidth}{!}{\begin{tabular}{cccccccccccccccccccccccccccc}
    \toprule
    \multirow{2}{*}{Method} 
    & \multicolumn{3}{c}{Mountain} 
    & \multicolumn{3}{c}{Fountain} 
    & \multicolumn{3}{c}{Corner} 
    & \multicolumn{3}{c}{Patio} 
    & \multicolumn{3}{c}{Spot} 
    & \multicolumn{3}{c}{Patio-High}\\
    & PSNR $\uparrow$ & SSIM $\uparrow$ & LPIPS $\downarrow$ 
    & PSNR $\uparrow$ & SSIM $\uparrow$ & LPIPS $\downarrow$ 
    & PSNR $\uparrow$ & SSIM $\uparrow$ & LPIPS $\downarrow$
    & PSNR $\uparrow$ & SSIM $\uparrow$ & LPIPS $\downarrow$ 
    & PSNR $\uparrow$ & SSIM $\uparrow$ & LPIPS $\downarrow$ 
    & PSNR $\uparrow$ & SSIM $\uparrow$ & LPIPS $\downarrow$ \\
    \midrule
    \midrule
    3DGS \cite{kerbl20233d}  
    & \metrictablethird{14.53} & 0.343 & 0.402 & 13.40 & 0.317 & 0.404 & 15.12 & 0.423 & 0.402 & 15.21 & 0.527 & 0.564 & 14.23 & 0.434 & 0.531 & 13.24 & 0.321 & 0.574 \\
    RobustSplats \cite{fu2025robustsplat}  
    & 6.053 & 0.063 & 0.775 & 10.98 & 0.274 & 0.658 & 15.27 & 0.428 & 0.489 & 9.686 & 0.241 & 0.693 & \metrictablethird{15.93} & 0.419 & 0.675 & 13.18 & 0.323 & 0.540 \\
    GS-W \cite{zhang2024gaussian}  
    & 13.47 & 0.336 & 0.471 & 11.17 & 0.177 & 0.618 & 14.34 & 0.535 & 0.481 & 13.47 & 0.336 & 0.471 & 14.34 & \metrictablesecond{0.504} & 0.481 & 13.47 & 0.336 & 0.571 \\
    DroneSplat \cite{tang2025dronesplat}  
    & 14.05 & 0.316 & 0.421 & 12.43 & 0.309 & 0.446 & \metrictablethird{16.68} & 0.521 & \metrictablethird{0.311} & 14.44 & 0.418 & \metrictablethird{0.433} & 14.48 & 0.350 & 0.543 & 12.65 & 0.274 & \metrictablesecond{0.542} \\
    WildGaussians \cite{kulhanek2024wildgaussians}  
    & 12.47 & \metrictablethird{0.555} & \metrictablethird{0.387} & \metrictablethird{13.34} & \metrictablethird{0.385} & \metrictablethird{0.527} & 15.56 & 0.514 & 0.354 & \metrictablethird{16.10} & \metrictablethird{0.573} & 0.445 & \metrictablethird{15.33} & 0.458 & \metrictablethird{0.503} & \metrictablethird{14.63} & \metrictablethird{0.431} & 0.599 \\
    Difix3D+ \cite{wu2025difix3d+}  
    & \metrictablesecond{15.21} & \metrictablesecond{0.606} & \metrictablesecond{0.334} & \metrictablesecond{16.70} & \metrictablesecond{0.593} & \metrictablesecond{0.299} & \metrictablesecond{18.03} & \metrictablesecond{0.597} & \metrictablesecond{0.286} & \metrictablesecond{17.45} & \metrictablesecond{0.659} & \metrictablesecond{0.431} & \metrictablesecond{17.48} & \metrictablethird{0.501} & \metrictablesecond{0.429} & \metrictablesecond{16.31} & \metrictablesecond{0.524} & \metrictablethird{0.546} \\
    Difix3D-W (Ours)  
    & \metrictablefirst{23.03} & \metrictablefirst{0.721} & \metrictablefirst{0.288} & \metrictablefirst{17.05} & \metrictablefirst{0.612} & \metrictablefirst{0.231} & \metrictablefirst{19.59} & \metrictablefirst{0.638} & \metrictablefirst{0.262} & \metrictablefirst{18.49} & \metrictablefirst{0.703} & \metrictablefirst{0.408} & \metrictablefirst{17.81} & \metrictablefirst{0.522} & \metrictablefirst{0.599} & \metrictablefirst{17.26} & \metrictablefirst{0.541} & \metrictablefirst{0.511} \\
    \bottomrule
    \end{tabular}}
    \caption{Quantitative results on the NeRF On-the-go dataset under a 9-view training setting. Performance is ranked by color from \metrictablethird{third} \metrictablesecond{to} \metrictablefirst{first}.}
    \label{tab:onthego_9view}
\end{table*}
\clearpage
{
    \small
    \bibliographystyle{ieeenat_fullname}
    \bibliography{main}
}

\end{document}